\documentclass{article}

\usepackage[main]{neurips_2026}

\usepackage[utf8]{inputenc} 
\usepackage[T1]{fontenc}    
\usepackage{hyperref}       
\usepackage{url}            
\usepackage{booktabs}       
\usepackage{amsfonts}       
\usepackage{nicefrac}       
\usepackage{microtype}      
\usepackage{xcolor}         
\usepackage{amsmath} 
 \usepackage{graphicx}
\usepackage{multirow} 

\title{\textsc{TRACER}: Verifiable Generative Provenance for Multimodal Tool-Using Agents}

\author{\bf
Bihui Yu$^{1,2}$,
Caijun Jia$^{1,2}$,
Jing Chi$^{4}$,
Xiaohan Liu$^{4}$,
Yining Wang$^{3}$,
He Bai$^{3}$,\\
\textbf{Yuchen Liu$^{3}$},
\textbf{Jingxuan Wei$^{1,2,4}$},
\textbf{Junnan Zhu$^{3}$}\thanks{Corresponding author.}\\[6pt]
\normalfont
$^1$ Shenyang Institute of Computing Technology, Chinese Academy of Sciences\\
$^2$ University of Chinese Academy of Sciences\\
$^3$ MAIS, Institute of Automation, Chinese Academy of Sciences\\
$^4$ Key Laboratory of Computing Power Network and Information Security, \\
Ministry of Education, Shandong Computer Science Center \\ (National Supercomputer Center in Jinan), \\
Qilu University of Technology (Shandong Academy of Sciences)\\[4pt]
}

\usepackage{graphicx}
\usepackage[most]{tcolorbox}
\usepackage{enumitem}
\usepackage{listings}
\usepackage{capt-of} 

\definecolor{promptcolor}{HTML}{EAF3FF}
\definecolor{promptcolorheader}{HTML}{8DB7E8}
\definecolor{codeback}{HTML}{F7FAFF}
\newtcolorbox{promptbox}[1][]{
  enhanced,
  breakable,
  top=0.3em,
  bottom=0.3em,
  left=0.5em,
  right=0.5em,
  toptitle=0.3em,
  bottomtitle=0.2em,
  boxsep=0pt,
  colframe=promptcolorheader,
  colback=promptcolor!50,
  boxrule=0.5pt,
  width=\columnwidth,
  title={\footnotesize #1}
}

\begin{document}

\maketitle
\begin{abstract}
Multimodal large language models (MLLMs) increasingly solve vision-centric tasks by calling external tools for visual inspection, OCR, retrieval, calculation, and multi-step reasoning.
Current tool-using agents usually expose the executed tool trajectory and the final answer, but they rarely specify which tool observation supports each generated claim.
We call this missing claim-level dependency structure the \emph{provenance gap}.
The gap makes tool use hard to verify and hard to optimize, because useful evidence, redundant exploration, and unsupported reasoning are mixed in the same trajectory.
We introduce \textsc{TRACER}, a framework for verifiable generative provenance in multimodal tool-using agents.
Instead of adding citations after generation, TRACER generates each answer sentence together with a structured provenance record that identifies the supporting tool turn, evidence unit, and semantic support relation.
Its relation space contains \textsc{Quotation}, \textsc{Compression}, and \textsc{Inference}, covering direct reuse, faithful condensation, and grounded derivation.
TRACER verifies each record through schema checking, tool-turn alignment, source authenticity, and relation rationality, and then converts verified provenance into traceability constraints and provenance-derived local credit for reinforcement learning.
We further construct \textsc{TRACE-Bench}, a benchmark for sentence-level provenance reconstruction from coarse multimodal tool trajectories.
On \textsc{TRACE-Bench}, simply adding tools often introduces noise.
With Qwen3-VL-8B-Instruct, TRACER reaches 78.23\% answer accuracy and 95.72\% summary accuracy, outperforming the strongest closed-source tool-augmented baseline by 23.80 percentage points.
Compared with tool-only supervised fine-tuning, it also reduces total test-set tool calls from 4{,}949 to 3{,}486.
These results show that reliable multimodal tool reasoning depends on provenance-aware use of observations, not on more tool calls alone.
\end{abstract}

\section{Introduction}
\label{sec:introduction}

MLLMs are increasingly used as tool-grounded agents for vision-centric tasks that require visual inspection, OCR, retrieval, calculation, and multi-step reasoning~\cite{caffagni2024revolution,liang2024survey,song2025bridge}.
Recent multimodal search and deep-research agents extend this paradigm by training models to acquire and integrate information across modalities and tools~\cite{jin2025search,wu2025mmsearch,geng2025webwatcher,huang2026vision}.
However, most tool-using MLLMs still provide only a trajectory-level view of reasoning: they record which tools were called and what final answer was produced.
They do not make explicit how individual answer claims depend on specific tool observations.
We refer to this missing claim-level dependency structure as the \emph{provenance gap}.

\begin{figure*}[t]
\setlength{\abovecaptionskip}{0.12cm}
\setlength{\belowcaptionskip}{-0.55cm}
    \centering
    \includegraphics[width=\linewidth]{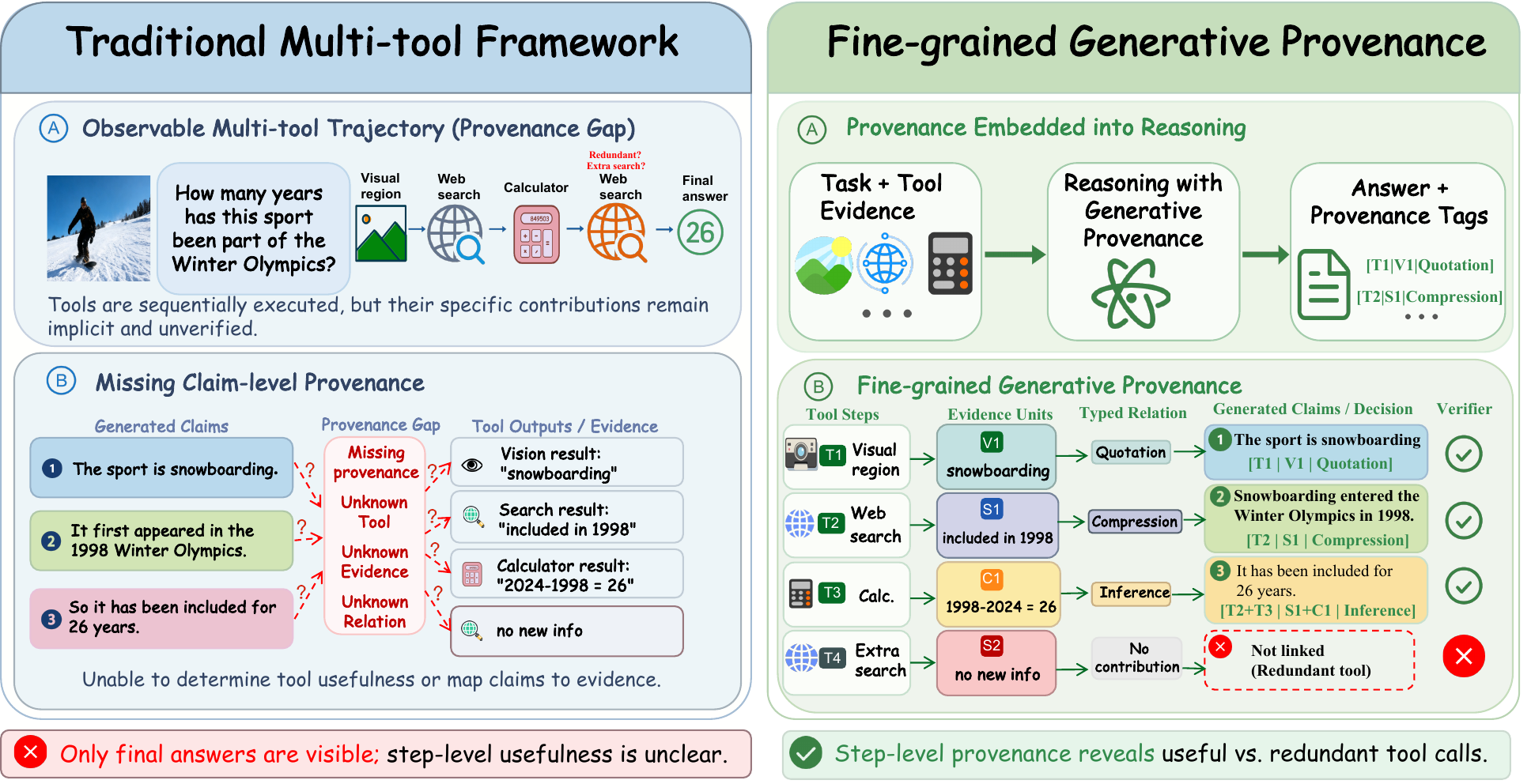}
    \caption{
    Overview of the provenance gap in multi-tool reasoning.
    Standard tool-using agents expose a tool trajectory and a final answer, but the claim-level dependency between tool observations and generated statements remains implicit.
    \textsc{TRACER} makes this dependency explicit by linking answer claims to tool turns, evidence units, and typed support relations, which separates provenance-bearing tool calls from redundant exploration.
    }
    \label{fig:motivation}
\end{figure*}

As shown in Figure~\ref{fig:motivation}, the provenance gap weakens both verification and learning.
A single trajectory can contain useful observations, failed or redundant calls, and unsupported reasoning steps.
If the final answer does not state which observations support which claims, a correct response may come from faithful tool use, parametric memory, accidental guessing, or ungrounded inference.
This ambiguity also creates a credit-assignment problem.
Answer-level or trajectory-level supervision cannot identify which tool turns actually contributed evidence, nor can it distinguish useful observations from noisy exploration.
The problem is especially acute in multimodal settings, where support may appear as image regions, OCR spans, retrieved snippets, visual descriptions, object counts, or computed values.
Reliable tool reasoning therefore requires an explicit provenance layer that makes generated claims traceable to the tool observations from which they are derived.

To bridge this gap, we introduce \textsc{TRACER}, a framework for verifiable generative provenance in multimodal tool-using agents.
TRACER treats provenance as part of generation rather than as a post-hoc explanation.
For each generated sentence, it emits a structured provenance record that specifies the supporting tool turn, the support unit, and the semantic relation between the tool observation and the claim.
The relation space contains \textsc{Quotation}, \textsc{Compression}, and \textsc{Inference}, which respectively capture direct reuse, faithful condensation, and grounded derivation.
Thus, a valid provenance record must identify not only \emph{where} a claim is grounded, but also \emph{how} the cited observation supports it.

TRACER further makes provenance an optimization interface.
Generated records are checked for structural validity, tool-turn alignment, source authenticity, and relation rationality.
Verified provenance is then used to identify provenance-bearing tool calls and assign local credit during group-relative reinforcement learning.
The resulting objective rewards correct and traceable answers, penalizes fabricated or unsupported provenance, and discourages unnecessary tool exploration.

To train and evaluate this capability, we construct \textsc{TRACE-Bench} for provenance-aware multimodal multi-tool reasoning.
Starting from coarse tool-interaction trajectories, \textsc{TRACE-Bench} reconstructs sentence-level reasoning paths with structured JSON provenance.
Each answer sentence is linked to a supporting tool turn, an evidence unit, and a typed relation.
The benchmark evaluates final answer accuracy, summary accuracy (SummAcc.), traceability accuracy, provenance F1, tool errors, and total tool-call usage, thereby measuring both task success and the faithfulness of claim-level grounding.

Experiments show that tool access alone does not reliably improve strong models.
Without provenance-aware training, additional tools can introduce noise and amplify planning errors.
In contrast, a Qwen3-VL-8B-Instruct instantiation of TRACER reaches 78.23\% accuracy and 95.72\% SummAcc., outperforming the strongest closed-source tool-augmented baseline by 23.80 percentage points.
Compared with tool-only supervised fine-tuning, TRACER improves accuracy while reducing total test-set tool calls from 4{,}949 to 3{,}486.
Ablations and qualitative analyses show that these gains come from provenance-aware use of tool observations rather than from longer trajectories or more sampled tool calls.

Overall, this paper makes three contributions.
(1) We formulate the \emph{provenance gap} in multimodal tool-using agents, where final claims lack an explicit dependency structure over intermediate tool observations.
(2) We introduce \textsc{TRACER}, a verifiable generative provenance framework that constructs sentence-level provenance during generation and converts verified provenance into local credit, traceability constraints, and efficient tool-use incentives.
(3) We construct \textsc{TRACE-Bench}, a benchmark for provenance-aware multimodal tool reasoning, and show that TRACER improves answer accuracy, provenance faithfulness, and tool-use efficiency.

\section{Related Work}
\paragraph{Multimodal tool-using agents.}
Recent multimodal agents move beyond single-turn perception toward long-horizon interaction, where models iteratively search, inspect, calculate, and integrate information across tools and modalities.
Search-R1~\cite{jin2025search} and MMSearch-R1~\cite{wu2025mmsearch} integrate external search into model reasoning, first in text-dominant settings and then in multimodal search over images and text.
WebWatcher~\cite{geng2025webwatcher} and Vision-DeepResearch~\cite{huang2026vision} study deep-research-style interaction in visually grounded environments.
DeepEyesV2~\cite{hong2025deepeyesv2}, Agent0-VL~\cite{liu2025agent0}, and ReAgent-V~\cite{zhou2025reagent} further improve tool-mediated reasoning, training stability, and video-oriented interaction.
Other approaches address unnecessary tool use, context growth, and scalable trajectory collection~\cite{yan2026act,du2026towards,peng2026mta,liu2026points}, while benchmarks stress multimodal search, sustained information integration, and workflow-level tool use~\cite{hong2025deepeyesv2,ning2025mc,zeng2026vision,su2026agentvista,dong2026epibench}.
These advances improve how agents acquire information and complete complex workflows.
However, they often leave the provenance structure of the final response implicit: a trajectory may contain the right page, visual observation, or computed value, while the answer still lacks a claim-level account of which observations warrant which statements.
TRACER targets this missing layer by requiring tool-grounded reasoning to produce verifiable generative provenance.

\paragraph{Provenance, citation, and process supervision.}
Citation and grounding methods aim to make model outputs more verifiable, but most of them attach references at a coarse granularity or after generation.
Recent text-provenance work provides finer-grained accounts of how generated text originates from sources.
TROVE~\cite{zhu2025trove} formulates provenance as sentence-level tracing with typed relations, including \textsc{Quotation}, \textsc{Compression}, and \textsc{Inference}.
GenProve~\cite{wei2026genprove} studies generation-time fine-grained provenance, where models generate answers together with structured provenance triples.
We adopt this typed view of provenance, but study a different setting.
Instead of tracing text answers to static source documents, TRACER links multimodal tool-grounded answers to tool turns, evidence units, and semantic support relations in dynamic interaction trajectories.
It also converts verified provenance into local credit for tool use, which is not addressed by text-only provenance generation.

Process supervision and agent-alignment methods provide denser feedback for intermediate behavior~\cite{choudhury2025process,xi2025agentprm,wang2025steca,wang2025spa,kazemnejad2024vineppo}.
Recent methods score progress, compare steps, propagate delayed feedback, or refine tool-integrated reinforcement learning~\cite{deng2024novice,tan2026hindsight,liang2026learning,ding2025empowering,wang2026enhancing,xue2025simpletir,guo2026e3}.
These methods supervise actions, progress, or outcomes, but they do not directly model claim-level provenance over concrete multimodal tool observations.
TRACER complements process supervision by making the evidential dependency structure explicit, verifiable, and usable for both evaluation and training.

\section{Method}
\label{sec:method}
TRACER addresses the provenance gap by making the dependency between answer claims and tool evidence explicit, verifiable, and optimizable.
Given a multimodal query, the model interacts with tools, generates sentence-level provenance for its final answer, and uses verified provenance to construct rewards and local credit signals for group-relative reinforcement learning.

\begin{figure}[t]
    \centering
    \includegraphics[width=1.0\textwidth]{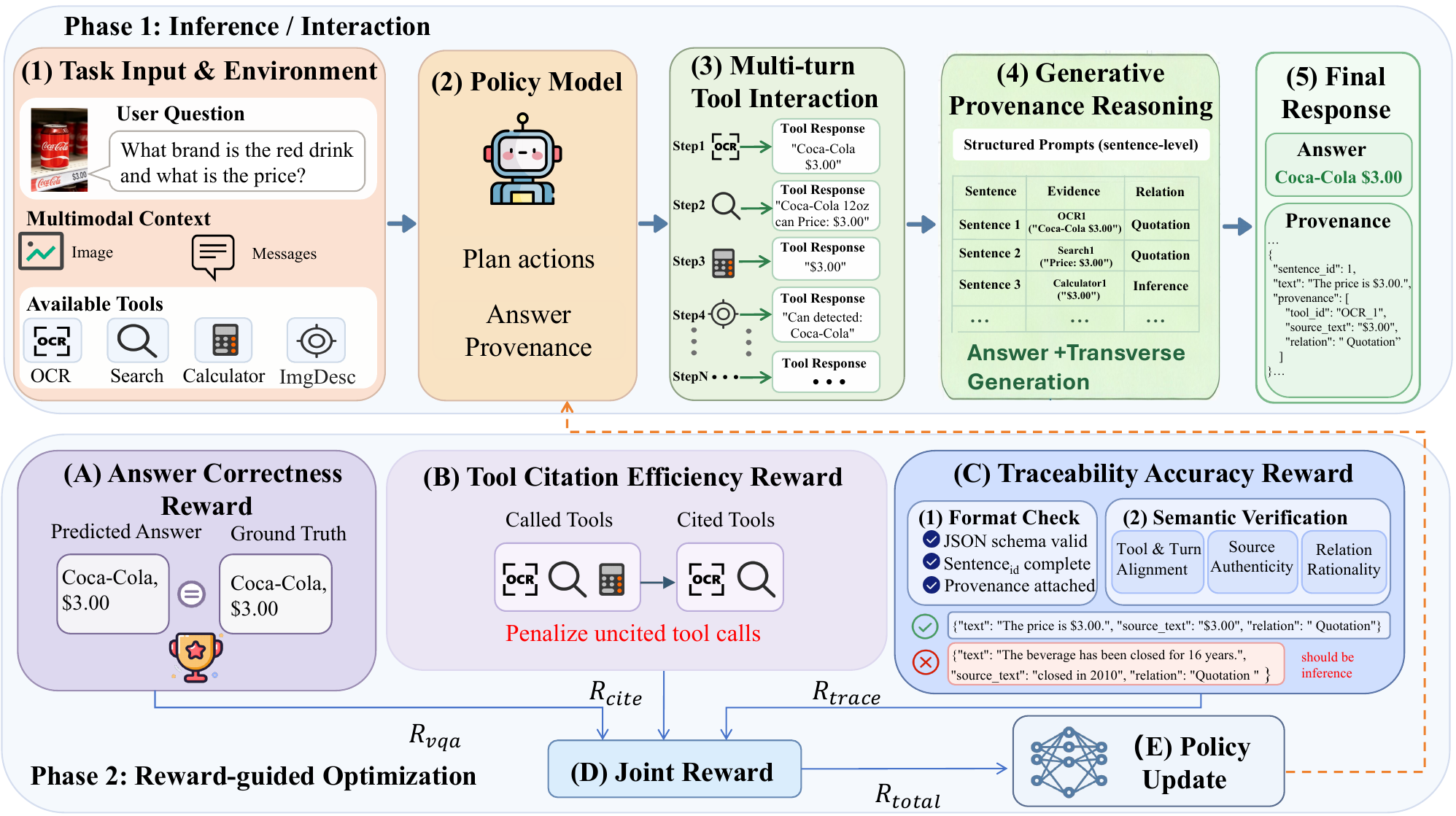}
    \caption{
    Illustration of \textsc{TRACER}.
    The policy model performs multi-turn tool-grounded reasoning, generates a final response with sentence-level provenance, and receives provenance-aware rewards for group-relative reinforcement learning.
    }
    \label{fig:tracer}
    \vspace{-1.0em}
\end{figure}

\subsection{Provenance-aware Multi-tool Reasoning}
\label{sec:method_formulation}

Each instance in \textsc{TRACE-Bench} consists of a question $q$, a multimodal context $\mathcal{M}$, an optional history $\mathcal{H}_0$, a tool set $\mathcal{U}$, and a ground-truth answer $y^\star$.
We denote the input as $x=(q,\mathcal{M},\mathcal{H}_0,\mathcal{U})$.
A policy model $\pi_\theta$ produces a tool-grounded trajectory $\tau=(a_1,e_1,\ldots,a_K,e_K)$, where $a_t=(u_t,\alpha_t)$ is the $t$-th tool query, $u_t\in\mathcal{U}$ is the selected tool, $\alpha_t$ is its argument, and $e_t$ is the returned evidence.
Each tool query has a unique turn identifier $\ell_t$, such as \texttt{OCR\_1} or \texttt{Calculator\_2}.

A standard tool-using agent generates only a final answer $y$ from $(x,\tau)$.
TRACER instead generates both an answer and a provenance structure:
$(y,\mathcal{G})\sim \pi_\theta(\cdot\mid x,\tau)$.
We decompose the answer into sentence-level claims $y=(s_1,\ldots,s_N)$.
For each sentence $s_i$, TRACER emits a provenance set
$\mathcal{P}_i=\{(\ell_{ij},z_{ij},r_{ij})\}_{j=1}^{m_i}$,
where $\ell_{ij}$ refers to a tool turn, $z_{ij}$ is the cited evidence span or localized multimodal evidence, and $r_{ij}$ specifies the support relation:
$r_{ij}\in\{\textsc{Quotation},\textsc{Compression},\textsc{Inference}\}$.
The complete provenance graph is $\mathcal{G}=\{(s_i,\mathcal{P}_i)\}_{i=1}^{N}$.
This formulation makes the evidential dependency between final claims and intermediate tool evidence explicit, which a trajectory-level correctness reward cannot provide.

\subsection{Generative Provenance and Provenance Rewards}
\label{sec:method_reward}

TRACER generates each sentence together with its provenance:
\begin{equation}
    p_\theta(y,\mathcal{G}\mid x,\tau)
    =
    \prod_{i=1}^{N}
    p_\theta(s_i,\mathcal{P}_i
    \mid x,\tau,s_{<i},\mathcal{P}_{<i}).
\end{equation}
The structured output follows a JSON schema that records the sentence text, sentence identifier, cited tool turn, source evidence, and relation type.

We verify provenance with a schema check and semantic traceability checks.
The schema check $C_{\mathrm{schema}}\in\{0,1\}$ verifies that the output is valid, each sentence is indexed, and each sentence has at least one provenance item.
For each provenance item, the traceability verifier returns
$v^{\mathrm{turn}}_{ij}$, $v^{\mathrm{src}}_{ij}$, and $v^{\mathrm{rel}}_{ij}\in\{0,1\}$,
which respectively check the existence of the cited tool turn, the authenticity of the cited source, and whether the declared relation supports the sentence.
The item-level validity is:
\begin{equation}
    v_{ij}
    =
    v^{\mathrm{turn}}_{ij}
    v^{\mathrm{src}}_{ij}
    v^{\mathrm{rel}}_{ij}.
\end{equation}

TRACER uses three rewards.
The answer reward is
\begin{equation}
    R_{\mathrm{vqa}}
    =
    \mathbb{I}\big[\mathrm{Eval}(y)=y^\star\big].
\end{equation}
Let $\mathcal{T}_{\mathrm{called}}=\{\ell_t\}_{t=1}^{K}$ be the called tool turns, and let $\mathcal{T}_{\mathrm{cited}}$ be the subset of tool turns appearing in at least one valid provenance item.
The citation-efficiency reward is
\begin{equation}
    R_{\mathrm{cite}}
    =
    \frac{
    |\mathcal{T}_{\mathrm{cited}}|
    }{
    \max(|\mathcal{T}_{\mathrm{called}}|,1)
    }.
\end{equation}
It penalizes tool calls that never support the final response.

The traceability reward is a hard faithfulness gate:
\begin{equation}
    R_{\mathrm{trace}}
    =
    C_{\mathrm{schema}}
    \prod_{i=1}^{N}
    \prod_{j=1}^{m_i}
    v_{ij}.
\end{equation}
It equals one only when every sentence-level provenance item passes all checks.
The final reward is
\begin{equation}
    R_{\mathrm{total}}
    =
    R_{\mathrm{vqa}}
    R_{\mathrm{trace}}
    \left(
    w_0 + w_{\mathrm{cite}} R_{\mathrm{cite}}
    \right),
\end{equation}
where $w_0$ preserves the reward for a correct and fully traceable answer, and $w_{\mathrm{cite}}$ encourages compact tool use.
This design treats traceability as a strict requirement rather than a soft bonus: an answer with fabricated or unsupported provenance receives zero training reward.

\subsection{Provenance-guided Group-relative Reinforcement}
\label{sec:method_optimization}

TRACER converts verified provenance into tool-turn credit.
For the $t$-th tool turn, the local credit is
\begin{equation}
    c_t
    =
    R_{\mathrm{vqa}}R_{\mathrm{trace}}
    \min\left(
    1,
    \sum_{i,j}
    \mathbb{I}[\ell_{ij}=\ell_t]v_{ij}
    \right).
\end{equation}
Thus, a tool turn receives credit only when it contributes verified evidence to a correct and fully traceable answer. For each input $x$, the old policy samples a group of $G$ rollouts:
\begin{equation}
    \{(\tau^{(g)},y^{(g)},\mathcal{G}^{(g)})\}_{g=1}^{G}
    \sim \pi_{\theta_{\mathrm{old}}}(\cdot\mid x).
\end{equation}
For token $n$ in rollout $g$, we define a provenance-aware group-normalized advantage:
\begin{equation}
    \widehat{A}^{(g)}_n
    =
    \operatorname{Norm}_{g\in[G]}
    \left(
    R_{\mathrm{total}}^{(g)}
    +
    \lambda
    \sum_t
    \mathbb{I}[n\in\mathcal{S}^{(g)}_t]c_t^{(g)}
    \right),
\end{equation}
where $\mathcal{S}^{(g)}_t$ denotes the token span that generates the $t$-th tool query, and $\lambda$ controls the strength of local provenance credit.
Non-tool tokens receive only the trajectory-level reward term.
This group-relative advantage does not require a learned value model. Let $\omega^{(g)}$ be the generated token sequence, $h^{(g)}_n$ be the prefix before token $n$, and
\begin{equation}
    \rho^{(g)}_n(\theta)
    =
    \frac{
    \pi_\theta(\omega^{(g)}_n\mid h^{(g)}_n)
    }{
    \pi_{\theta_{\mathrm{old}}}(\omega^{(g)}_n\mid h^{(g)}_n)
    }.
\end{equation}
We optimize a clipped per-token objective:
\begin{equation}
\begin{aligned}
    \ell^{(g)}_n(\theta)
    =
    &\min\Big(
    \rho^{(g)}_n\widehat{A}^{(g)}_n,\,
    \operatorname{clip}(\rho^{(g)}_n,1-\epsilon,1+\epsilon)
    \widehat{A}^{(g)}_n
    \Big) \\
    &-
    \beta
    D_{\mathrm{KL}}
    \big(
    \pi_\theta(\cdot\mid h^{(g)}_n)
    \,\|\, 
    \pi_{\mathrm{ref}}(\cdot\mid h^{(g)}_n)
    \big).
\end{aligned}
\end{equation}
The final optimization objective is
\begin{equation}
    \mathcal{J}_{\textsc{TRACER}}(\theta)
    =
    \mathbb{E}_{x}
    \mathbb{E}_{\{\omega^{(g)}\}}
    \left[
    \frac{1}{G}
    \sum_{g=1}^{G}
    \frac{1}{|\omega^{(g)}|}
    \sum_{n=1}^{|\omega^{(g)}|}
    \ell^{(g)}_n(\theta)
    \right].
\end{equation}
The policy maximizes $\mathcal{J}_{\textsc{TRACER}}$.
The group-normalized reward compares multiple reasoning trajectories for the same input, while provenance credit identifies which tool queries actually support the final answer.
As a result, TRACER aligns optimization with the evidential structure of multi-tool reasoning rather than assigning the same trajectory-level reward to all intermediate decisions.

\section{Dataset}
\label{sec:dataset}

\begin{figure*}[t]
    \centering
    \includegraphics[width=1.0\textwidth]{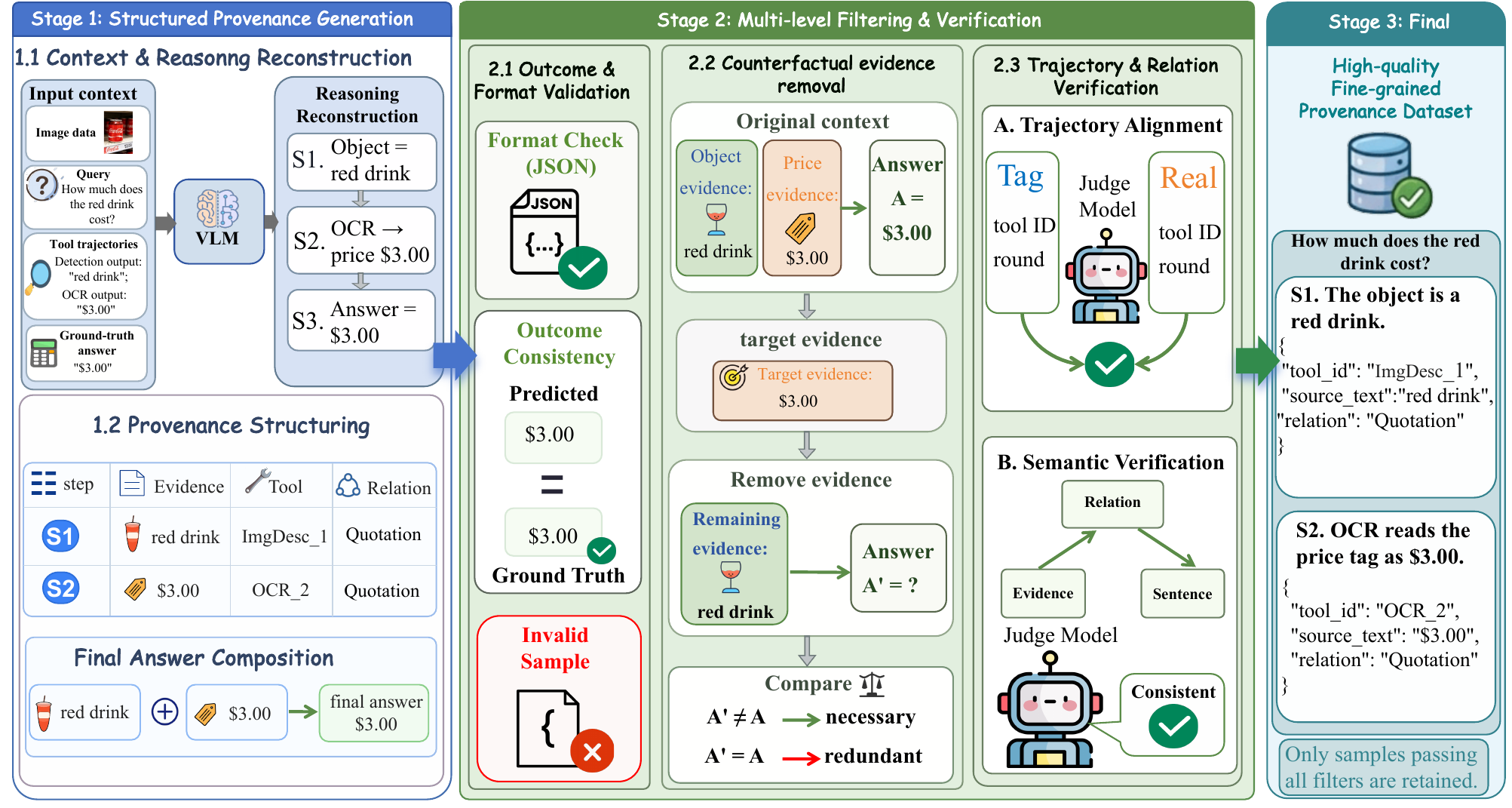}
    \caption{
    Overview of the \textsc{TRACE-Bench} construction pipeline.
    Coarse tool-interaction trajectories are transformed into sentence-level provenance records and retained only after outcome, format, counterfactual, and provenance verification.
    }
    \label{fig:dataset_pipeline}
    \vspace{-1.0em}
\end{figure*}

\textsc{TRACE-Bench} evaluates whether tool-augmented multimodal agents can produce correct answers together with verifiable claim-level provenance.
We build it on ToolVQA~\cite{yin2025toolvqa}, whose trajectories contain multimodal inputs, questions, tool calls, tool observations, and reference answers.
The original trajectories do not specify which observation supports each final claim.
We therefore transform coarse multi-turn trajectories into concise reasoning responses with structured provenance records that identify the exact tool observation supporting each answer sentence.

\subsection{Provenance Construction}
\label{sec:source_provenance_construction}

Each source instance consists of a question $q$, a multimodal context $\mathcal{M}$, a tool set $\mathcal{U}$, a ground-truth answer $y^\star$, and an ordered tool trajectory $\tau$.
We construct a provenance-aware output $(y,\mathcal{G})$, where $y$ is decomposed into sentence-level claims and $\mathcal{G}$ records the supporting evidence for each claim.
For each sentence, the provenance specifies the cited tool turn, the localized evidence span, and the semantic relation, chosen from \textsc{Quotation}, \textsc{Compression}, and \textsc{Inference}.
We instantiate this process with a VLM-based generator prompted with the full task context and the reference answer.
The generator produces a coherent response and a JSON provenance record for every sentence, converting each original trajectory into an explicit claim, i.e., evidence graph.
Details on the tool inventory and relation taxonomy are provided in Appendix~\ref{app:dataset_statistics_and_analysis}, and the generation and verification prompts are shown in Figures~\ref{fig:multi_tool_vqa_provenance_prompt} and~\ref{fig:provenance_evaluation_prompt}.

\subsection{Multi-stage Filtering}
\label{sec:dataset_filtering}

Because generated provenance can contain answer inconsistencies, malformed JSON, unsupported citations, or incorrect relation labels, we apply a multi-stage filtering pipeline.
First, we validate answer consistency against the reference answer and check the JSON schema with a rule-based parser.
Second, we perform counterfactual evidence removal: for each cited evidence unit, we remove or mask that unit while preserving the rest of the trajectory and then re-evaluate the answer and provenance.
If the answer and provenance remain unchanged, the citation is likely redundant, recoverable from other observations, or unsupported by the claimed source, so the record is filtered.
If removing the evidence changes the answer or provenance, we treat the record as evidence-sensitive under the verifier and retain it for the next stage.
Finally, an LLM-as-a-judge verifier checks trajectory alignment, source authenticity, and relation rationality, discarding records with nonexistent tool turns, untraceable evidence, or implausible evidence--relation--claim triples.
Additional details and common failure modes are discussed in Appendix~\ref{app:dataset_statistics_and_analysis}.

\subsection{Split Construction and Statistics}
\label{sec:split_construction}

After filtering, we retain 14{,}183 high-quality provenance-aware trajectories from 23{,}655 candidates, corresponding to a 59.96\% retention rate.
We preserve the official ToolVQA held-out test split of 2{,}550 examples and partition the remaining examples into 10{,}470 training and 1{,}163 validation instances.
The final responses are compact, with an average of 2.4 reasoning sentences and 2.56 anchored provenance records per response.
Comprehensive statistics on dataset scale, tool-call composition, text length, and relation distributions are presented in Appendix~\ref{app:dataset_statistics}.

\section{Experimental}
\label{sec:Experimental}

\subsection{Main Results}
\label{sec:main_results}

\begin{table*}[t]
\centering
\caption{
Main results on \textsc{TRACE-Bench}.
Acc. denotes final task accuracy.
Tool Call and Tool Error are total counts on the held-out test set.
SummAcc. measures the quality of tool-trajectory summarization for the target question.
Traceability Acc. and Prov. F1 evaluate generated claim-level provenance and are applicable only to models that output structured provenance records.
}
\label{tab:main_results}
\resizebox{\textwidth}{!}{
\begin{tabular}{lrrrrrr}
\toprule
Model Setting
& Acc. (\%)
& Tool Call $\downarrow$
& Tool Error $\downarrow$
& SummAcc. (\%)
& Traceability Acc. (\%)
& Prov. F1 (\%) \\
\midrule
\multicolumn{7}{c}{\textit{Closed-source LFMs}} \\
\midrule
ChatGPT-4o-latest~\cite{hurst2024gpt} & 38.29 & -- & -- & -- & -- & -- \\
ChatGPT-4o-latest + tool & 34.96 & 2003 & 1804 & 56.10 & -- & -- \\
Claude-3-5-sonnet~\cite{Anthropic2024ClaudeSonnet3_5} & 30.33 & -- & -- & -- & -- & -- \\
Claude-3-5-sonnet + tool & 30.52 & 585 & 524 & 75.52 & -- & -- \\
Gemini-2.5-Pro~\cite{google2025gemini25} & 27.35 & -- & -- & -- & -- & -- \\
Gemini-2.5-Pro + tool & 54.43 & 1569 & 438 & 70.27 & -- & -- \\
\midrule
\multicolumn{7}{c}{\textit{Open-source LFMs}} \\
\midrule
Qwen2-VL-7B-Instruct~\cite{wang2024qwen2} & 3.64 & -- & -- & -- & -- & -- \\
Qwen2-VL-7B-Instruct + tool & 5.86 & 1373 & 1051 & 38.18 & -- & -- \\
LLaVA-v1.5-7B VLM~\cite{liu2024improved} & 8.57 & -- & -- & -- & -- & -- \\
LLaVA-v1.5-7B VLM + tool & 1.17 & 9684 & 9684 & 0.01 & -- & -- \\
Qwen2-VL-2B-Instruct~\cite{wang2024qwen2} & 7.71 & -- & -- & -- & -- & -- \\
Qwen2-VL-2B-Instruct + tool & 2.10 & 4067 & 3915 & 21.82 & -- & -- \\
Tuned LLaVA-7B~\cite{yin2025toolvqa} & 7.21 & -- & -- & -- & -- & -- \\
Tuned LLaVA-7B + tool & 18.80 & 4311 & 617 & 30.91 & -- & -- \\
InternVL3.5-7B~\cite{wang2025internvl3} & 12.27 & -- & -- & -- & -- & -- \\
InternVL3.5-7B + tool & 13.68 & 249 & 10 & 48.39 & -- & -- \\
MiMo-VL-7B-RL~\cite{xiaomi2025mimo} & 22.66 & -- & -- & -- & -- & -- \\
MiMo-VL-7B-RL + tool & 37.09 & 8491 & 1513 & 57.31 & -- & -- \\
Qwen3-VL-8B-Instruct~\cite{bai2025qwen3} & 23.52 & -- & -- & -- & -- & -- \\
Qwen3-VL-8B-Instruct + tool & 26.62 & 5580 & 3730 & 72.07 & -- & -- \\
Qwen3-VL-8B-Instruct + tool-SFT & 70.94 & 4949 & 590 & 84.39 & -- & -- \\
\midrule
TRACER-SFT & 73.56 & 4034 & 654 & 93.29 & 77.53 & 70.70 \\
\textbf{TRACER-RL} & \textbf{78.23} & \textbf{3486} & \textbf{573} & \textbf{95.72} & \textbf{93.61} & \textbf{90.52} \\
\bottomrule
\end{tabular}
}
\end{table*}

Table~\ref{tab:main_results} reports the main experimental results, and the detailed experimental configurations are provided in Appendix~\ref{app:experimental_details}.
In addition to final task accuracy and tool-use statistics, we include provenance quality metrics in the main table, since the central goal of \textsc{TRACER} is not only to improve answer accuracy, but also to make generated answers verifiably grounded in external tool observations.
Traceability Acc. and provenance F1 are reported for models that generate structured provenance records.
Overall, the results show that simply equipping models with external tools does not consistently improve task performance.
For example, ChatGPT-4o-latest reaches 38.29\% accuracy without tools, but its tool-augmented variant drops to 34.96\% while producing 2{,}003 tool calls and 1{,}804 tool errors.
Several open-source models show the same pattern: they issue many tool calls but obtain limited or negative accuracy gains.
LLaVA-v1.5-7B VLM + tool, for instance, produces 9{,}684 calls, all counted as tool errors under our evaluation protocol, and reaches only 0.01\% SummAcc.
These results indicate that tool use is not a simple interface extension; it requires planning, evidence selection, observation integration, and claim-level grounding.

Supervised training on tool trajectories substantially improves tool interaction.
Qwen3-VL-8B-Instruct + tool-SFT reaches 70.94\% accuracy, outperforming the untuned tool-augmented counterpart by 44.32 percentage points.
However, tool-only SFT still lacks explicit supervision over which observations support which final claims, and it continues to use a relatively large number of tools.
Adding generative provenance supervision improves both task performance and grounding: \textsc{TRACER-SFT} reaches 73.56\% accuracy, 93.29\% SummAcc., 77.53\% Traceability Acc., and 70.70\% provenance F1.
After reinforcement learning, \textsc{TRACER-RL} achieves the best results, with 78.23\% accuracy, 95.72\% SummAcc., 93.61\% Traceability Acc., and 90.52\% provenance F1.
It outperforms the strongest closed-source tool-augmented baseline, Gemini-2.5-Pro + tool, by 23.80 percentage points.
Compared with Qwen3-VL-8B-Instruct + tool-SFT, it also reduces total test-set tool calls from 4{,}949 to 3{,}486, a 29.56\% reduction.
Thus, the improvement comes from using fewer and more relevant observations, not from calling more tools.

\subsection{Ablation Study}
\label{sec:ablation_study}

We ablate the two provenance-aware reward components to test whether the gains of \textsc{TRACER-RL} come from the proposed provenance mechanism.
The citation-efficiency reward $R_{\mathrm{cite}}$ is designed to reduce tool calls that do not support the final response.
The traceability reward $R_{\mathrm{trace}}$ enforces faithful alignment between generated provenance records and actual tool observations.
We evaluate the effect of removing each reward on answer accuracy, traceability, provenance precision and recall, and tool-use efficiency.

\begin{table*}[ht]
\centering
\caption{
Ablation results of \textsc{TRACER-RL}.
Provenance-aware rewards improve answer accuracy, traceability, provenance quality, and tool-use efficiency.
}
\label{tab:rl_ablation}
\resizebox{\textwidth}{!}{
\begin{tabular}{lrrrrrrrr}
\toprule
Model Setting
& Acc. (\%)
& Tool Call
& Tool Error
& SummAcc. (\%)
& Traceability Acc. (\%)
& Prov. Precision (\%)
& Prov. Recall (\%)
& Prov. F1 (\%) \\
\midrule
TRACER-SFT
& 73.56
& 4034
& 654
& 93.29
& 77.53
& 68.42
& 73.15
& 70.70 \\
TRACER-RL
& \textbf{78.23}
& \textbf{3486}
& \textbf{573}
& \textbf{95.72}
& \textbf{93.61}
& \textbf{89.24}
& \textbf{91.85}
& \textbf{90.52} \\
TRACER-RL w/o $R_{\mathrm{cite}}$
& 75.13
& 4251
& 615
& 93.86
& 87.28
& 74.20
& 85.12
& 79.28 \\
TRACER-RL w/o $R_{\mathrm{trace}}$
& 74.58
& 3839
& 593
& 92.14
& 91.67
& 82.15
& 83.40
& 82.77 \\
\bottomrule
\end{tabular}
}
\end{table*}

As shown in Table~\ref{tab:rl_ablation}, the full TRACER-RL model consistently outperforms TRACER-SFT across both task-level and provenance-level metrics.
Accuracy increases from 73.56\% to 78.23\%, Traceability Acc. improves from 77.53\% to 93.61\%, and provenance F1 increases substantially from 70.70\% to 90.52\%.
This confirms that the RL stage does not simply improve answer selection, but teaches the model to generate answers whose claims are more faithfully grounded in the original tool trajectory.
The two reward ablations reveal complementary failure modes.
When $R_{\mathrm{cite}}$ is removed, the number of tool calls rises from 3486 to 4251, while provenance precision drops sharply from 89.24\% to 74.20\%.
Although provenance recall remains relatively high at 85.12\%, the precision degradation indicates that many additional tool observations do not provide reliable support for the final claims.
This suggests that, without citation-efficiency constraints, the model tends to over-explore and becomes less selective about which observations should be treated as provenance-bearing evidence.
When $R_{\mathrm{trace}}$ is removed, the model produces fewer tool calls than the variant without $R_{\mathrm{cite}}$, but its accuracy drops to 74.58\%, SummAcc. drops to 92.14\%, and provenance F1 decreases to 82.77\%.
This shows that reducing redundant tool calls alone is insufficient for faithful reasoning, since the model may still generate plausible-looking provenance records that do not correctly match the gold evidence units or support relations.

Therefore, $R_{\mathrm{cite}}$ mainly improves tool-use efficiency by suppressing redundant exploration and increasing the density of useful evidence, while $R_{\mathrm{trace}}$ improves provenance faithfulness by enforcing valid alignment among claims, evidence units, tool turns, and relation types.
Only when both rewards are used together does the model achieve high accuracy, low tool-call count, high traceability, and strong provenance F1.
These results provide direct evidence that TRACER improves multimodal tool reasoning by learning a verifiable provenance structure, rather than merely benefiting from additional RL training or more sampled trajectories.

\subsection{Case Study}
\label{sec:case_study}
\begin{figure*}[t]
    \centering
    \includegraphics[width=1.0\textwidth]{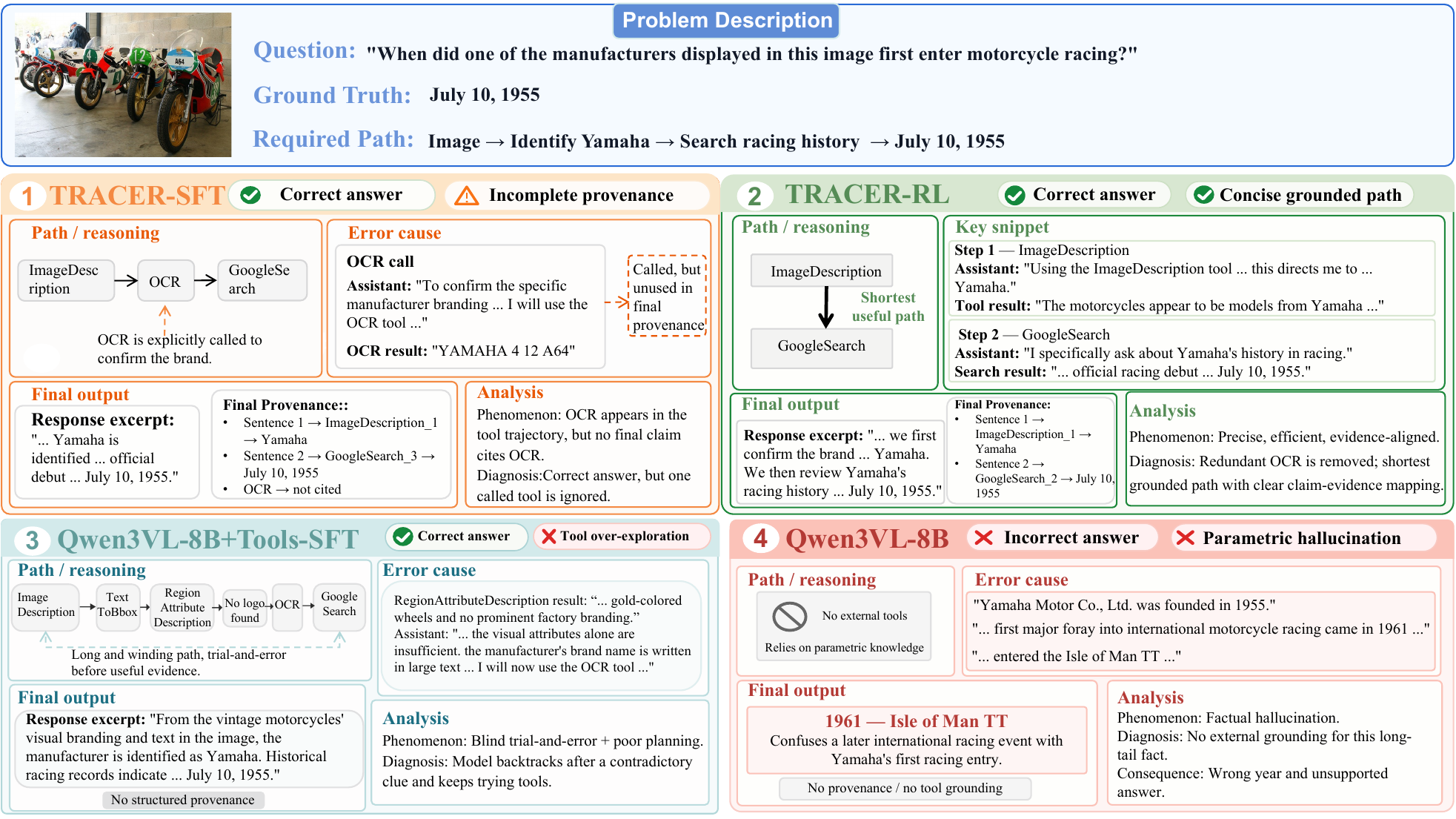}
    \vspace{-2.0em}
    \caption{
    Comparison of reasoning trajectories. Unlike traditional methods that treat all tool calls as necessary evidence, \textbf{TRACER} exposes the actual dependency between observations and claims. Generative provenance reveals redundant steps (e.g., OCR in SFT) and enables \textbf{TRACER-RL} to learn more efficient, evidence-directed tool-use trajectories. 
    }
\end{figure*}

We present a representative case to qualitatively examine how generative provenance affects multi-tool visual reasoning. The question asks when one of the manufacturers shown in the image first entered motorcycle racing. Solving it requires identifying the visual manufacturer and retrieving a fine-grained historical fact, making it difficult to answer reliably from parametric knowledge alone.

The tool-free model predicts an incorrect year, suggesting that memorized knowledge is insufficient for this long-tail multimodal factual query. The tool-augmented model without provenance training can access external tools, but follows an inefficient trajectory by invoking \texttt{TextToBbox}, \texttt{RegionAttributeDescription}, OCR, and \texttt{GoogleSearch}. Several early localization and regional-description calls fail to provide useful evidence, indicating that tool access alone does not ensure effective evidence acquisition. TRACER-SFT produces the correct answer with \texttt{ImageDescription}, OCR, and \texttt{GoogleSearch}. More importantly, its generated provenance exposes the actual dependency structure behind the answer: the manufacturer identification is supported by \texttt{ImageDescription\_1}, while the racing-entry date is supported by \texttt{GoogleSearch\_3}. Although OCR is called, none of the final claims cite its output. Thus, a conventional tool trajectory would make OCR appear to be part of the reasoning chain, whereas generative provenance reveals it as a redundant exploration step. This illustrates the provenance gap: standard trajectories record which tools are called, but do not reveal which observations actually support each generated claim. TRACER-RL further removes the redundant OCR step and directly uses \texttt{ImageDescription} for visual identification and \texttt{GoogleSearch} for factual retrieval. This shorter trajectory still yields the correct answer, showing that reinforcement learning based on provenance signals does not merely encourage more tool use, but instead promotes compact and evidence-directed tool use. By penalizing unused citations and assigning credit to useful tool observations, TRACER-RL learns to suppress low-value exploration while preserving the evidence needed for answer generation. 

Overall, this case supports our main claim: reliable multi-tool visual reasoning requires not only access to external tools, but also explicit modeling of the dependency between tool observations and final claims. Generative provenance makes this dependency observable, enabling both better interpretability and more effective tool-use optimization.

\section{Conclusion}
\label{sec:conclusion}

We introduce \textsc{TRACER}, a framework for verifiable generative provenance in multimodal tool-using agents.
TRACER addresses the provenance gap by pairing each generated sentence with structured provenance that specifies the supporting tool turn, evidence unit, and semantic support relation.
This design turns provenance from a post-hoc explanation into a generation-time interface that can be verified and optimized.
By using verified provenance to distinguish provenance-bearing tool calls from redundant exploration, TRACER enables finer-grained credit assignment and more efficient tool use.
We also construct \textsc{TRACE-Bench}, a benchmark for evaluating provenance-aware multimodal tool reasoning from coarse interaction trajectories.
Experiments show that tool access alone can introduce noise, while provenance-aware training improves answer accuracy, summary accuracy, traceability, provenance F1, and tool-call efficiency.
These results suggest that reliable multimodal tool reasoning should be driven by faithful and verifiable use of supporting observations, rather than by more tool calls alone.

\bibliography{ref}

@inproceedings{jin2025search,
title={Search-R1: Training {LLM}s to Reason and Leverage Search Engines with Reinforcement Learning},
author={Bowen Jin and Hansi Zeng and Zhenrui Yue and Jinsung Yoon and Sercan O Arik and Dong Wang and Hamed Zamani and Jiawei Han},
booktitle={Second Conference on Language Modeling (COLM)},
year={2025},
url={https://openreview.net/forum?id=Rwhi91ideu}
}

@article{wu2025mmsearch,
  title={Mmsearch-r1: Incentivizing lmms to search},
  author={Wu, Jinming and Deng, Zihao and Li, Wei and Liu, Yiding and You, Bo and Li, Bo and Ma, Zejun and Liu, Ziwei},
  journal={arXiv preprint arXiv:2506.20670},
  year={2025}
}

@inproceedings{geng2025webwatcher,
  title={WebWatcher: Breaking New Frontiers of Vision-Language Deep Research Agent},
  author={Xinyu Geng and Peng Xia and Zhen Zhang and Xinyu Wang and Qiuchen Wang and Ruixue Ding and Chenxi Wang and Jialong Wu and Kuan Li and Yida Zhao and Huifeng Yin and Yong Jiang and Pengjun Xie and Fei Huang and Huaxiu Yao and Yi R. Fung and Jingren Zhou},
  booktitle={The Fourteenth International Conference on Learning Representations (ICLR)},
  year={2026},
  url={https://openreview.net/forum?id=8jsaazdAb3}
}

@inproceedings{hong2025deepeyesv2,
  title={DeepEyesV2: Toward Agentic Multimodal Model},
  author={Jack Hong and Chenxiao Zhao and ChengLIn Zhu and Weiheng Lu and Guohai Xu and XingYu},
  booktitle={The Fourteenth International Conference on Learning Representations (ICLR)},
  year={2026},
  url={https://openreview.net/forum?id=yDKawwfJ5O}
}

@inproceedings{zhou2025reagent,
  title={ReAgent-V: A Reward-Driven Multi-Agent Framework for Video Understanding},
  author={Yiyang Zhou and Yangfan He and Yaofeng Su and Siwei Han and Joel Jang and Gedas Bertasius and Mohit Bansal and Huaxiu Yao},
  booktitle={The Thirty-ninth Annual Conference on Neural Information Processing Systems (NeurIPS)},
  year={2026},
  url={https://openreview.net/forum?id=D1Iw4Unvfc}
}

@inproceedings{ning2025mc,
  title={Search-{MM}: Benchmarking Multimodal Agentic {RAG} with Structured Reasoning Chains},
  author={Xuying Ning and Dongqi Fu and Tianxin Wei and Mengting Ai and Jiaru Zou and Ting-Wei Li and Jingrui He},
  booktitle={NeurIPS 2025 Workshop on Evaluating the Evolving LLM Lifecycle: Benchmarks, Emergent Abilities, and Scaling},
  year={2025},
  url={https://openreview.net/forum?id=S2zaYgT7Ic}
}

@article{huang2026vision,
  title={Vision-deepresearch: Incentivizing deepresearch capability in multimodal large language models},
  author={Huang, Wenxuan and Zeng, Yu and Wang, Qiuchen and Fang, Zhen and Cao, Shaosheng and Chu, Zheng and Yin, Qingyu and Chen, Shuang and Yin, Zhenfei and Chen, Lin and others},
  journal={arXiv preprint arXiv:2601.22060},
  year={2026}
}

@article{zeng2026vision,
  title={Vision-deepresearch benchmark: Rethinking visual and textual search for multimodal large language models},
  author={Zeng, Yu and Huang, Wenxuan and Fang, Zhen and Chen, Shuang and Shen, Yufan and Cai, Yishuo and Wang, Xiaoman and Yin, Zhenfei and Chen, Lin and Chen, Zehui and others},
  journal={arXiv preprint arXiv:2602.02185},
  year={2026}
}

@article{liu2025agent0,
  title={Agent0-vl: Exploring self-evolving agent for tool-integrated vision-language reasoning},
  author={Liu, Jiaqi and Xiong, Kaiwen and Xia, Peng and Zhou, Yiyang and Ji, Haonian and Feng, Lu and Han, Siwei and Ding, Mingyu and Yao, Huaxiu},
  journal={arXiv preprint arXiv:2511.19900},
  year={2025}
}

@article{yan2026act,
  title={Act Wisely: Cultivating Meta-Cognitive Tool Use in Agentic Multimodal Models},
  author={Yan, Shilin and Tong, Jintao and Xue, Hongwei and Tang, Xiaojun and Wang, Yangyang and Shi, Kunyu and Zhang, Guannan and Li, Ruixuan and Zou, Yixiong},
  journal={arXiv preprint arXiv:2604.08545},
  year={2026}
}

@article{du2026towards,
  title={Towards Long-horizon Agentic Multimodal Search},
  author={Du, Yifan and Liu, Zikang and Peng, Jinbiao and Wu, Jie and Li, Junyi and Li, Jinyang and Zhao, Wayne Xin and Wen, Ji-Rong},
  journal={arXiv preprint arXiv:2604.12890},
  year={2026}
}

@article{peng2026mta,
  title={MTA-Agent: An Open Recipe for Multimodal Deep Search Agents},
  author={Peng, Xiangyu and Qin, Can and Yan, An and Yang, Xinyi and Chen, Zeyuan and Xu, Ran and Wu, Chien-Sheng},
  journal={arXiv preprint arXiv:2604.06376},
  year={2026}
}

@article{liu2026points,
  title={POINTS-Seeker: Towards Training a Multimodal Agentic Search Model from Scratch},
  author={Liu, Yikun and Liu, Yuan and Tian, Le and Zhou, Xiao and Yao, Jiangchao and Wang, Yanfeng and Xie, Weidi},
  journal={arXiv preprint arXiv:2604.14029},
  year={2026}
}

@article{su2026agentvista,
  title={Agentvista: Evaluating multimodal agents in ultra-challenging realistic visual scenarios},
  author={Su, Zhaochen and Gao, Jincheng and Guo, Hangyu and Liu, Zhenhua and Zhang, Lueyang and Geng, Xinyu and Huang, Shijue and Xia, Peng and Jiang, Guanyu and Wang, Cheng and others},
  journal={arXiv preprint arXiv:2602.23166},
  year={2026}
}

@article{dong2026epibench,
  title={EpiBench: Benchmarking Multi-turn Research Workflows for Multimodal Agents},
  author={Dong, Xuan and Zheng, Huanyang and Niu, Tianhao and Han, Zhe and Li, Pengzhan and Liu, Bofei and Liu, Zhengyang and Li, Guancheng and Zhu, Qingfu and Che, Wanxiang},
  journal={arXiv preprint arXiv:2604.05557},
  year={2026}
}

@article{choudhury2025process,
  title={Process reward models for llm agents: Practical framework and directions},
  author={Choudhury, Sanjiban},
  journal={arXiv preprint arXiv:2502.10325},
  year={2025}
}

@inproceedings{wang2025steca,
    title = "{ST}e{C}a: Step-level Trajectory Calibration for {LLM} Agent Learning",
    author = "Wang, Hanlin  and
      Wang, Jian  and
      Leong, Chak Tou  and
      Li, Wenjie",
    booktitle = "Findings of the Association for Computational Linguistics: ACL 2025",
    year = "2025",
    url = "https://aclanthology.org/2025.findings-acl.604/",
    pages = "11597--11614",
}

@article{wang2025spa,
  title={Spa-rl: Reinforcing llm agents via stepwise progress attribution},
  author={Wang, Hanlin and Leong, Chak Tou and Wang, Jiashuo and Wang, Jian and Li, Wenjie},
  journal={arXiv preprint arXiv:2505.20732},
  year={2025}
}

@article{deng2024novice,
  title={From novice to expert: Llm agent policy optimization via step-wise reinforcement learning},
  author={Deng, Zhirui and Dou, Zhicheng and Zhu, Yutao and Wen, Ji-Rong and Xiong, Ruibin and Wang, Mang and Chen, Weipeng},
  journal={arXiv preprint arXiv:2411.03817},
  year={2024}
}

@inproceedings{kazemnejad2024vineppo,
  title={Vine{PPO}: Refining Credit Assignment in {RL} Training of {LLM}s},
  author={Amirhossein Kazemnejad and Milad Aghajohari and Eva Portelance and Alessandro Sordoni and Siva Reddy and Aaron Courville and Nicolas Le Roux},
  booktitle={Forty-second International Conference on Machine Learning (ICML)},
  year={2025},
  url={https://openreview.net/forum?id=Myx2kJFzAn}
}

@inproceedings{xue2025simpletir,
  title={Simple{TIR}: End-to-End Reinforcement Learning for Multi-Turn Tool-Integrated Reasoning},
  author={Zhenghai Xue and Longtao Zheng and Qian Liu and Yingru Li and Xiaosen Zheng and Zejun MA and Bo An},
  booktitle={The Fourteenth International Conference on Learning Representations (ICLR)},
  year={2026},
  url={https://openreview.net/forum?id=EplNy91Xqh}
}

@inproceedings{xi2025agentprm,
  author = {Xi, Zhiheng and Liao, Chenyang and Li, Guanyu and Zhang, Zhihao and Chen, Wenxiang and Wang, Binghai and Jin, Senjie and Zhou, Yuhao and Guan, Jian and Wu, Wei and Ji, Tao and Gui, Tao and Zhang, Qi and Huang, Xuanjing},
  title = {AgentPRM: Process Reward Models for LLM Agents via Step-Wise Promise and Progress},
  year = {2026},
  url = {https://doi.org/10.1145/3774904.3792551},
  booktitle = {Proceedings of the ACM Web Conference 2026 (WWW)},
  pages = {4184–4195},
}

@article{tan2026hindsight,
  title={Hindsight Credit Assignment for Long-Horizon LLM Agents},
  author={Tan, Hui-Ze and Yang, Xiao-Wen and Chen, Hao and Shao, Jie-Jing and Wen, Yi and Shen, Yuteng and Luo, Weihong and Du, Xiku and Guo, Lan-Zhe and Li, Yu-Feng},
  journal={arXiv preprint arXiv:2603.08754},
  year={2026}
}

@article{liang2026learning,
  title={Learning from the Irrecoverable: Error-Localized Policy Optimization for Tool-Integrated LLM Reasoning},
  author={Liang, Qiao and Zhu, Yuke and Ge, Chao and Yang, Lei and Shen, Ying and Zheng, Bo and Guo, Sheng},
  journal={arXiv preprint arXiv:2602.09598},
  year={2026}
}

@article{ding2025empowering,
  title={Empowering Multi-Turn Tool-Integrated Reasoning with Group Turn Policy Optimization},
  author={Ding, Yifeng and Le, Hung and Han, Songyang and Ruan, Kangrui and Jin, Zhenghui and Kumar, Varun and Wang, Zijian and Deoras, Anoop},
  journal={arXiv preprint arXiv:2511.14846},
  year={2025}
}

@article{guo2026e3,
  title={E3-TIR: Enhanced Experience Exploitation for Tool-Integrated Reasoning},
  author={Guo, Weiyang and Shi, Zesheng and Zhao, Liye and Ma, Jiayuan and Zhu, Zeen and He, Junxian and Zhang, Min and Li, Jing},
  journal={arXiv preprint arXiv:2604.09455},
  year={2026}
}

@article{wang2026enhancing,
  title={Enhancing LLM-based Search Agents via Contribution Weighted Group Relative Policy Optimization},
  author={Wang, Junzhe and Xi, Zhiheng and Luo, Hao and Dou, Shihan and Gui, Tao and Zhang, Qi and others},
  journal={arXiv preprint arXiv:2604.14267},
  year={2026}
}

@inproceedings{caffagni2024revolution,
    title = "The Revolution of Multimodal Large Language Models: A Survey",
    author = "Caffagni, Davide  and
      Cocchi, Federico  and
      Barsellotti, Luca  and
      Moratelli, Nicholas  and
      Sarto, Sara  and
      Baraldi, Lorenzo  and
      Baraldi, Lorenzo  and
      Cornia, Marcella  and
      Cucchiara, Rita",
    booktitle = "Findings of the Association for Computational Linguistics: ACL 2024",
    year = "2024",
    url = "https://aclanthology.org/2024.findings-acl.807/",
    pages = "13590--13618",
}

@inproceedings{liang2024survey,
  title={A survey of multimodel large language models},
  author={Liang, Zijing and Xu, Yanjie and Hong, Yifan and Shang, Penghui and Wang, Qi and Fu, Qiang and Liu, Ke},
  booktitle={Proceedings of the 3rd international conference on computer, artificial intelligence and control engineering},
  pages={405--409},
  year={2024}
}

@article{song2025bridge,
  title={How to bridge the gap between modalities: Survey on multimodal large language model},
  author={Song, Shezheng and Li, Xiaopeng and Li, Shasha and Zhao, Shan and Yu, Jie and Ma, Jun and Mao, Xiaoguang and Zhang, Weimin and Wang, Meng},
  journal={IEEE Transactions on Knowledge and Data Engineering (TKDE)},
  volume={37},
  number={9},
  pages={5311--5329},
  year={2025},
  publisher={IEEE}
}

@article{wei2026genprove,
  title={GenProve: Learning to Generate Text with Fine-Grained Provenance},
  author={Wei, Jingxuan and Wang, Xingyue and Liao, Yanghaoyu and Dong, Jie and Liu, Yuchen and Jia, Caijun and Yu, Bihui and Zhu, Junnan},
  journal={arXiv preprint arXiv:2601.04932},
  year={2026}
}

@inproceedings{zhu2025trove,
    title = "{TROVE}: A Challenge for Fine-Grained Text Provenance via Source Sentence Tracing and Relationship Classification",
    author = "Zhu, Junnan  and
      Xiao, Min  and
      Wang, Yining  and
      Zhai, Feifei  and
      Zhou, Yu  and
      Zong, Chengqing",
    booktitle = "Proceedings of the 63rd Annual Meeting of the Association for Computational Linguistics (ACL)",
    year = "2025",
    url = "https://aclanthology.org/2025.acl-long.577/",
    pages = "11755--11771",
}

@inproceedings{yin2025toolvqa,
  title={Toolvqa: A dataset for multi-step reasoning vqa with external tools},
  author={Yin, Shaofeng and Lei, Ting and Liu, Yang},
  booktitle={Proceedings of the IEEE/CVF International Conference on Computer Vision},
  pages={4424--4433},
  year={2025}
}

@article{bai2025qwen3,
  title={Qwen3-vl technical report},
  author={Bai, Shuai and Cai, Yuxuan and Chen, Ruizhe and Chen, Keqin and Chen, Xionghui and Cheng, Zesen and Deng, Lianghao and Ding, Wei and Gao, Chang and Ge, Chunjiang and others},
  journal={arXiv preprint arXiv:2511.21631},
  year={2025}
}

@article{hurst2024gpt,
  title={Gpt-4o system card},
  author={Hurst, Aaron and Lerer, Adam and Goucher, Adam P and Perelman, Adam and Ramesh, Aditya and Clark, Aidan and Ostrow, AJ and Welihinda, Akila and Hayes, Alan and Radford, Alec and others},
  journal={arXiv preprint arXiv:2410.21276},
  year={2024}
}

@techreport{Anthropic2024ClaudeSonnet3_5,
  title        = {Claude 3.5 Sonnet System Card},
  author       = {Anthropic},
  institution  = {Anthropic PBC},
  year         = {2024},
  note         = {Official technical report describing Claude 3.5 Sonnet capabilities and safety benchmarks. Available at: \url{https://www.anthropic.com/news/claude-3-5-sonnet}}
}

@article{google2025gemini25,
  title={Gemini 2.5: Pushing the Frontier with Advanced Reasoning, Multimodality, Long Context, and Next Generation Agentic Capabilities},
  author={Comanici, Gheorghe and Bieber, Eric and Schaekermann, Mike and Pasupat, Ice and Sachdeva, Noveen and Dhillon, Inderjit and Blistein, Marcel and Ram, Ori and Zhang, Dan and others},
  journal={arXiv preprint arXiv:2507.06261},
  year={2025},
}

@article{wang2024qwen2,
  title={Qwen2-vl: Enhancing vision-language model's perception of the world at any resolution},
  author={Wang, Peng and Bai, Shuai and Tan, Sinan and Wang, Shijie and Fan, Zhihao and Bai, Jinze and Chen, Keqin and Liu, Xuejing and Wang, Jialin and Ge, Wenbin and others},
  journal={arXiv preprint arXiv:2409.12191},
  year={2024}
}

@inproceedings{liu2024improved,
  title={Improved baselines with visual instruction tuning},
  author={Liu, Haotian and Li, Chunyuan and Li, Yuheng and Lee, Yong Jae},
  booktitle={Proceedings of the IEEE/CVF conference on computer vision and pattern recognition},
  pages={26296--26306},
  year={2024}
}

@article{wang2025internvl3,
  title={Internvl3. 5: Advancing open-source multimodal models in versatility, reasoning, and efficiency},
  author={Wang, Weiyun and Gao, Zhangwei and Gu, Lixin and Pu, Hengjun and Cui, Long and Wei, Xingguang and Liu, Zhaoyang and Jing, Linglin and Ye, Shenglong and Shao, Jie and others},
  journal={arXiv preprint arXiv:2508.18265},
  year={2025}
}

@article{xiaomi2025mimo,
  title={MiMo: Unlocking the Reasoning Potential of Language Model--From Pretraining to Posttraining},
  author={Xiaomi, LLM and Xia, Bingquan and Shen, Bowen and Zhu, Dawei and Zhang, Di and Wang, Gang and Zhang, Hailin and Liu, Huaqiu and Xiao, Jiebao and Dong, Jinhao and others},
  journal={arXiv preprint arXiv:2505.07608},
  year={2025}
}
\bibliographystyle{unsrt}


\appendix
\section{Limitations and Broader Impacts}
\label{app:Limitations}

\paragraph{Limitations.}
TRACER improves verifiable generative provenance for multimodal tool-using agents, but several limitations remain.
First, \textsc{TRACE-Bench} relies on LLM-as-a-judge verification during data filtering and traceability evaluation.
Although we combine schema checks, counterfactual evidence removal, trajectory alignment, and relation-rationality checks, evaluator bias and prompt sensitivity may still introduce noise.
Second, the benchmark is built on ToolVQA and currently covers seven tools for global visual description, OCR, localization, region description, counting, retrieval, and calculation.
Additional tools, modalities, and task domains are needed to test whether the same provenance interface generalizes to broader agentic workflows.
Third, our trainable models are based on an 8B-scale MLLM.
The scaling behavior of generative provenance, local tool credit, and citation-efficiency rewards for substantially larger models remains an important direction for future work.
Finally, provenance verification checks whether cited evidence supports generated claims, but it does not prove that the model internally used the evidence causally.
We therefore interpret provenance as a verifiable external dependency structure, not as a complete mechanistic explanation of the model's internal computation.

\paragraph{Broader impacts.}
The main positive impact of TRACER is improved accountability for tool-using multimodal agents.
By requiring each claim to cite authentic tool observations and a rational support relation, the framework can help users audit generated answers, identify unsupported claims, and reduce redundant tool usage.
Potential risks include over-trusting provenance records when verifiers are imperfect, using generated provenance to create misleading but plausible explanations, or applying tool-using agents in sensitive domains without domain-specific validation.
Mitigations include reporting verifier limitations, retaining the original tool trajectories for audit, releasing data documentation with source and license information, and avoiding deployment in high-stakes settings without independent verification.

\section{Experimental Details}
\label{app:experimental_details}

\subsection{Experimental Setup}
\label{app:experimental_setup}

We evaluate on \textsc{TRACE-Bench} using 10,470 training, 1,163 validation, and 2,550 held-out test examples in accordance with the ToolVQA protocol~\cite{yin2025toolvqa}. Operating under a tool-interactive environment equipped with visual perception, OCR, grounding, retrieval, and computation capabilities, models iteratively gather observations to produce final answers alongside structured claim-level provenance. All trainable models are initialized from Qwen3-VL-8B-Instruct~\cite{bai2025qwen3}. We compare our provenance-aware \textsc{TRACER-SFT} and reward-optimized \textsc{TRACER-RL} against proprietary multimodal models, representative open-source baselines, and a standard tool-SFT baseline designed to isolate the impact of explicit provenance optimization. Model performance is comprehensively assessed across task-solving ability via answer and summary accuracy, provenance quality via traceability accuracy and provenance F1, and efficiency via the average number of tool calls per example.

\subsection{Training Details}
\label{app:training_details}

All trainable models are initialized from Qwen3-VL-8B-Instruct~\cite{bai2025qwen3}.
During validation and test inference, we use deterministic decoding with temperature set to 0.

\paragraph{Supervised fine-tuning.}
In the supervised fine-tuning stage, we perform full-parameter updates across the vision encoder, multi-modal projector, and language model backbone.
The maximum input length is set to 8,192 tokens.
We use a per-device batch size of 2 with gradient accumulation over 4 steps, a learning rate of $5 \times 10^{-6}$, and a 5\% warmup ratio.
Training is conducted for 2 epochs using bfloat16 precision and DeepSpeed ZeRO-2 optimization across 8 NVIDIA A100 GPUs.
The SFT objective trains the model to follow the tool-interaction format and generate sentence-level provenance records for final answers.

\paragraph{Reinforcement learning.}
Following SFT, we apply reinforcement learning to further improve task performance and provenance faithfulness.
The RL model is initialized from the SFT checkpoint and optimized for 200 steps.
For each prompt, we generate a rollout group of 8 candidate outputs and adopt rejection sampling to construct optimization signals.
We further curate high-entropy training examples whose sampled candidates show substantial variance between correct and incorrect responses, focusing optimization on uncertain cases.

\paragraph{Reinforcement learning.}
The RL reward follows the provenance-aware reward defined in Section~\ref{sec:method_reward}.
It consists of three components: answer correctness, tool citation efficiency, and traceability accuracy.
Answer correctness and traceability accuracy are used as hard gating signals.
If the predicted answer is incorrect or the generated provenance fails the traceability verification, the overall reward is set to zero.
Otherwise, the reward is computed as
\begin{equation}
    R_{\mathrm{total}}
    =
    R_{\mathrm{vqa}}
    R_{\mathrm{trace}}
    \left(
    w_0 + w_{\mathrm{cite}} R_{\mathrm{cite}}
    \right),
\end{equation}
where we set $w_0=1.0$ and $w_{\mathrm{cite}}=0.5$ in all experiments.
Here, $R_{\mathrm{vqa}}$ indicates whether the final answer is correct, $R_{\mathrm{trace}}$ indicates whether all sentence-level provenance items pass schema, source-alignment, and relation-rationality checks, and $R_{\mathrm{cite}}$ measures the fraction of called tool turns that are actually cited by valid provenance.
This design rewards correct and fully traceable answers while encouraging compact tool use and penalizing unnecessary or uncited tool calls.

We further use provenance-derived local credit to guide optimization over tool-query tokens.
Following Section~\ref{sec:method_optimization}, the local credit coefficient is set to $\lambda=0.3$.
The clipped group-relative objective uses a clipping threshold of $\epsilon=0.2$.
A KL penalty against the SFT reference model is applied during RL training, with coefficient $\beta=0.02$.
The reward weights, local credit coefficient, clipping threshold, and KL coefficient are selected on the validation set and kept fixed for all RL experiments.
All training stages are implemented with \texttt{torchrun} and distributed across 8 NVIDIA A100 GPUs.


\subsection{Baseline Details}
\label{app:baseline_details}

We compare \textsc{TRACER} with a broad set of closed-source and open-source multimodal large foundation models.

\paragraph{Closed-source models.}
We include ChatGPT-4o-latest~\cite{hurst2024gpt}, Claude-3-5-Sonnet~\cite{Anthropic2024ClaudeSonnet3_5}, and Gemini-2.5-Pro~\cite{google2025gemini25}.
For each closed-source model, we evaluate both the direct-answer setting and the tool-augmented setting.
In the direct-answer setting, the model directly predicts the answer from the input image and question.
In the tool-augmented setting, the model is allowed to interact with the same tool environment before producing the final answer.

\paragraph{Open-source models.}
We evaluate representative open-source vision-language models, including Qwen2-VL-7B-Instruct~\cite{wang2024qwen2}, LLaVA-v1.5-7B~\cite{liu2024improved}, Qwen2-VL-2B-Instruct~\cite{wang2024qwen2}, Tuned LLaVA-7B~\cite{yin2025toolvqa}, InternVL3.5-7B~\cite{wang2025internvl3}, MiMo-VL-7B-RL~\cite{xiaomi2025mimo}, and Qwen3-VL-8B-Instruct~\cite{bai2025qwen3}.
Whenever applicable, each model is evaluated in both direct-answer and tool-augmented modes.
These baselines test whether existing open-source multimodal models can effectively plan tool calls, integrate multi-turn observations, and summarize intermediate evidence into the final answer.

\paragraph{Tool-use and provenance-aware baselines.}
We include Qwen3-VL-8B-Instruct+tool-SFT as a strong supervised tool-use baseline.
This model is trained with the same base checkpoint and tool-interaction format as our method, but does not explicitly optimize claim-level generative provenance.
We further report \textsc{TRACER-SFT}, which uses provenance-aware supervised fine-tuning, and \textsc{TRACER-RL}, which additionally applies reward-guided optimization with answer correctness, tool citation efficiency, and traceability accuracy rewards.
For models that do not produce structured provenance records, Traceability Acc. and provenance F1 are not applicable.

\subsection{Metric Details}
\label{app:metric_details}

We evaluate models using both task-solving and provenance-oriented metrics.

\paragraph{Answer accuracy.}
Answer accuracy (Acc.) measures whether the model produces the correct final answer for the original VQA task.
It reflects end-to-end performance in the tool-interactive setting.

\paragraph{Summary accuracy.}
Summary accuracy (SummAcc.) measures whether the model can infer the correct final answer from a given tool trajectory without making additional tool calls.
This metric isolates the model's ability to integrate and summarize already observed tool evidence.

\paragraph{Traceability accuracy.}
Traceability Acc. is computed using an LLM-as-a-judge verifier, where we use Gemini-2.5-Pro as the judge model.
Given the generated provenance and the original tool trajectory, the verifier checks whether each cited tool identifier corresponds to a real tool call, whether the cited evidence can be traced back to the selected tool observation, and whether the predicted provenance relation is semantically reasonable.
An example is counted as traceable only when its generated claim-level provenance passes all these checks.

\paragraph{Provenance precision, recall, and F1.}
Provenance precision, recall, and F1 are computed against the reference provenance annotations in the test set.
We treat each provenance annotation as an evidence--claim link with a relation type.
Precision measures the proportion of predicted provenance links that are correct, recall measures the proportion of reference provenance links recovered by the model, and F1 is their harmonic mean.
These metrics evaluate whether the model can generate verifiable and complete claim-level provenance rather than merely producing a correct final answer.

\paragraph{Tool-call statistics.}
We additionally report the average number of tool calls per example to characterize tool-use behavior.
This metric helps distinguish models that solve tasks through efficient evidence acquisition from those that rely on redundant or poorly grounded tool exploration.

\section{Additional Dataset Details and Statistics}
\label{app:dataset_statistics_and_analysis}

\subsection{Comprehensive Dataset Statistics}
\label{app:dataset_statistics}

Table~\ref{tab:dataset_summary} summarizes the overall dataset scale, tool-call composition, text length statistics, and provenance annotation distribution. Furthermore, Figure~\ref{fig:dataset_statistics} provides a compact statistical overview of the final benchmark, illustrating the data split, tool-call distribution, and provenance-relation distribution.

\begin{table*}[h]
\centering
\begin{minipage}{0.42\textwidth}
\centering
\small
\setlength{\tabcolsep}{4pt}
\begin{tabular}{l r}
\toprule
\textbf{Statistic} & \textbf{Value} \\
\midrule
\multicolumn{2}{l}{\textit{Dataset Overview}} \\
Total examples & 14,183 \\
Total tool calls & 40,491 \\
\midrule
\multicolumn{2}{l}{\textit{Tool-call Distribution}} \\
ImageDescription & 12,594 \\
GoogleSearch & 9,666 \\
OCR & 5,801 \\
Calculator & 3,281 \\
TextToBbox & 3,094 \\
RegionAttributeDescription & 3,094 \\
CountGivenObject & 2,961 \\
\midrule
\multicolumn{2}{l}{\textit{Length Statistics}} \\
Average query length & 15.46 \\
Average response length & 78.43 \\
\midrule
\multicolumn{2}{l}{\textit{Provenance Statistics}} \\
Total records & 36,431 \\
Avg records per example & 2.57 \\
Avg records per sentence & 1.07 \\
\midrule
\multicolumn{2}{l}{\textit{Relation Distribution}} \\
\textsc{Quotation} & 24,239 \\
\textsc{Inference} & 8,364 \\
\textsc{Compression} & 3,828 \\
\bottomrule
\end{tabular}
\caption{Summary of the \textsc{TRACE-Bench} dataset.}
\label{tab:dataset_summary}
\end{minipage}
\hfill 
\begin{minipage}{0.54\textwidth} 
\centering
\includegraphics[width=\textwidth]{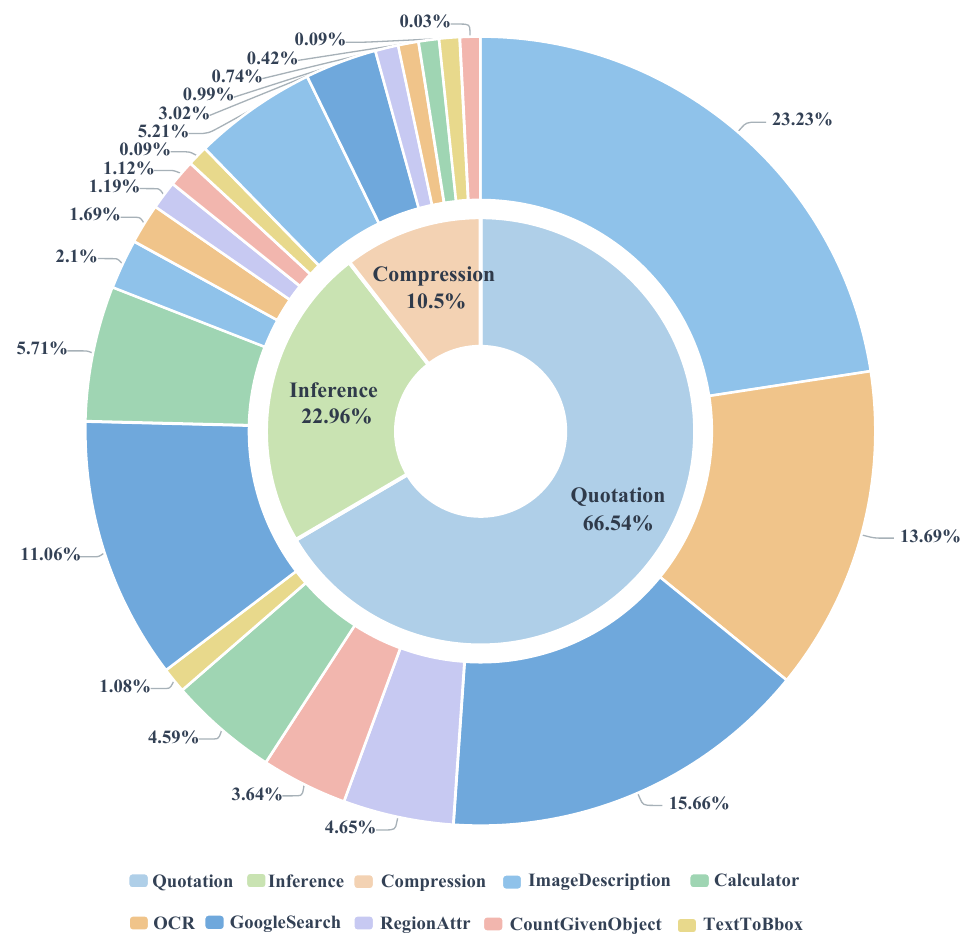}
\captionof{figure}{Statistical overview of \textsc{TRACE-Bench}. The figure summarizes the data split, tool-call distribution, and provenance-relation distribution.}
\label{fig:dataset_statistics}
\end{minipage}
\end{table*}

\subsection{Detailed Cross-domain Affinity}
\label{app:cross_domain_affinity}

As summarized in Table~\ref{tab:dataset_summary}, the dataset categorizes evidence into quotation at 66.5\%, inference at 22.9\%, and compression relations at 10.5\%, revealing a distinct cognitive division of labor. Further breakdown reveals distinct tool affinities. Among the 24,239 quotation relations, ImageDescription with 8,463 instances and OCR with 4,987 instances maintain absolute dominance. This aligns with the intuition of multimodal interaction: when processing complex images, visual models tend to prioritize quoting the extracted geometric and textual attributes of the scene as foundational factual anchors for reasoning. 

Within the 8,364 inference relations, GoogleSearch with 4,029 instances and Calculator with 2,080 instances exhibit exceptionally strong task affinity. Singular visual quotation is often insufficient for resolving complex visual question answering problems. The model must leverage external retrieval to introduce background information of searched entities or input quoted numerical values into a calculator for symbolic deduction, thereby completing an evidence-based deep logical leap. Lastly, the compression relation is primarily concentrated in ImageDescription with 1,899 instances and GoogleSearch with 1,101 instances. Because the observation states returned by these tools often contain substantial redundant text, the model spontaneously exhibits the ability to distill crucial state information through summarization and noise reduction mechanisms.

\subsection{Empirical Characteristics and Provenance Failures}
\label{app:empirical_characteristics}

\textsc{TRACE-Bench} emphasizes verifiable generative provenance rather than merely adding structured metadata to existing trajectories. The benchmark requires a model to identify provenance-bearing observations, distinguish them from redundant tool calls, and assign rational support relations to generated claims. This design is especially important for multimodal tool-use agents, where a correct final answer may still be produced from unsupported reasoning, irrelevant observations, or redundant exploration. 

The filtered examples reveal several common provenance failures. These include malformed provenance schemas, missing sentence-level evidence, invalid tool-turn references, evidence units that cannot be aligned to the cited observation, relation labels inconsistent with the evidence--claim transformation, and evidence-insensitive citations under counterfactual removal. Such failures correspond to realistic errors in tool-using agents. For instance, a model may call OCR but never use the OCR result in any final claim, or cite a retrieved snippet while actually relying on a previous visual description. Overall, \textsc{TRACE-Bench} provides an empirical testbed for studying provenance-aware optimization, local credit assignment, and efficient tool use in multimodal agents.

\subsection{Detailed Multi-stage Filtering Pipeline}
\label{app:filtering_details}

As briefly introduced in Section~\ref{sec:dataset_filtering}, because provenance annotations are generated from complex, long-context multimodal tool trajectories, they may inevitably contain answer-level inconsistencies, malformed JSON structures, unsupported evidence citations, or incorrect relation labels. To mitigate these issues and retain only high-confidence provenance-aware examples, we design and apply a rigorous three-stage filtering process.

\paragraph{Stage 1: Outcome Consistency and Schema Validation.}
The first stage serves as a foundational filter for correctness and structural integrity. We first verify whether the reconstructed final answer is semantically consistent with the reference answer $A_{\mathrm{gt}}$. Samples whose final conclusion contradicts the reference answer are strictly removed, as their provenance annotations might explain an incorrect reasoning path rather than the intended solution. Following outcome validation, we utilize a rule-based parser to validate the JSON provenance schema. Samples exhibiting parsing errors, malformed JSON structures, missing sentence-level provenance, invalid relation labels, nonexistent tool identifiers, or incomplete evidence fields are automatically discarded.

\paragraph{Stage 2: Counterfactual Evidence Removal.}
The second stage performs a rigorous evidence sensitivity check. For a claimed evidence unit $e_{ij}$, we construct a counterfactual context in which this specific evidence unit is removed or masked, while the remainder of the tool trajectory is perfectly preserved. The verifier then re-evaluates whether the generated answer and its supporting provenance remain stable under this modified context. If the prediction or the provenance structure becomes unstable after removing the cited evidence, we treat the record as exhibiting behavioral reliance under the verifier. Conversely, if the answer remains unchanged, it suggests that the cited evidence might be redundant, recoverable from another observation, or already embedded within the model's parametric knowledge. Therefore, rather than interpreting counterfactual removal as absolute causal proof of necessity, we utilize it as a conservative filtering signal to identify explicitly evidence-sensitive provenance records.

\paragraph{Stage 3: LLM-as-a-Judge Provenance Verification.}
Finally, an LLM-as-a-judge verifier is deployed to perform a deep semantic check on trajectory alignment and relation rationality. Trajectory alignment strictly requires that every cited tool identifier corresponds to an actual tool call within the original trajectory, and that every cited evidence unit is authentically traceable to the selected observation. This effectively prevents the model from hallucinating citations or misattributing evidence. Relation rationality evaluates whether the declared relation type logically matches the semantic operation executed between the cited evidence and the generated claim. Records containing unsupported evidence, incorrect tool alignment, or implausible evidence--relation--claim triples are permanently removed.

\subsection{Tool Inventory}
\label{app:tool_inventory}

\begin{table}[ht]
\centering
\small
\setlength{\tabcolsep}{4pt}
\begin{tabular}{p{0.24\linewidth} p{0.43\linewidth} p{0.23\linewidth}}
\toprule
\textbf{Tool} & \textbf{Function} & \textbf{Observation Type} \\
\midrule
ImageDescription 
& Produces a global description of the input image, including salient objects, scene layout, visual attributes, and visible entities. 
& Global visual description \\

OCR 
& Extracts text visible in the image. 
& Recognized text \\

TextToBbox 
& Grounds a textual object or phrase to its corresponding image region. 
& Bounding box or grounded region \\

RegionAttributeDescription 
& Describes attributes of a specified image region, such as color, shape, texture, or local semantics. 
& Region-level visual description \\

CountGivenObject 
& Counts instances of a specified object category in the image. 
& Object count \\

GoogleSearch 
& Retrieves external textual knowledge for open-domain entity or fact verification. 
& Retrieved textual snippet \\

Calculator 
& Performs deterministic arithmetic or symbolic computation. 
& Computed value \\
\bottomrule
\end{tabular}
\caption{
Tool inventory in \textsc{TRACE-Bench}.
The retained tools cover global visual perception, localized inspection, OCR, counting, retrieval, and deterministic computation.
}
\label{tab:tool_inventory}
\end{table}

\begin{table*}[ht]
\centering
\small
\setlength{\tabcolsep}{5pt}
\begin{tabular}{p{0.16\linewidth} p{0.48\linewidth} p{0.28\linewidth}}
\toprule
\textbf{Relation} & \textbf{Definition} & \textbf{Example} \\
\midrule
\textsc{Quotation} &
The claim directly reuses or reports information explicitly stated in the tool observation, with minimal semantic transformation. &
OCR returns ``Yamaha'', and the claim states that the visible text reads ``Yamaha''. \\
\textsc{Compression} &
The claim faithfully condenses a longer or redundant observation into a shorter statement without introducing new reasoning. &
An image description lists multiple visual details, and the claim summarizes that the image shows a Yamaha-branded motorcycle. \\
\textsc{Inference} &
The claim is derived by combining, comparing, or computing over one or more evidence units, while remaining grounded in the cited observations. &
A search result gives the first racing date and the image identifies the manufacturer; the claim answers when that manufacturer first entered motorcycle racing. \\
\bottomrule
\end{tabular}
\caption{
Relation taxonomy used in \textsc{TRACE-Bench}.
The taxonomy distinguishes direct reuse, faithful condensation, and grounded derivation.
}
\label{tab:app_relation_taxonomy}
\vspace{-1.0em}
\end{table*}

Table~\ref{tab:tool_inventory} summarizes the seven tools retained from the source trajectories in \textsc{TRACE-Bench}. These tools cover visual perception, localized image inspection, textual extraction, external knowledge retrieval, counting, and deterministic computation. For each tool, we report its functionality and the typical form of its observation returned to the agent.

\subsection{Provenance Relation Taxonomy}
\label{app:relation_taxonomy}

Table~\ref{tab:app_relation_taxonomy} defines the three relation types used in \textsc{TRACE-Bench}. The taxonomy describes how a cited evidence unit supports a generated claim, rather than simply identifying the source of the evidence.

\section{Provenance Generator Prompt}
\vspace{-1.2em}
\begin{promptbox}[Multi-Tool VQA Provenance Generator Prompt]
\sloppy\setlength{\emergencystretch}{1em}

You are a ``Multi-Tool VQA Provenance Answer Generator''.

\vspace{2pt}
\textbf{1. Input Data}

You will receive structured input data containing:
\begin{itemize}[leftmargin=*, itemsep=1pt]
  \item \texttt{image\_path}
  \item \texttt{context}: A sequential multi-tool invocation chain; each step contains \texttt{name}, \texttt{thought}, \texttt{input}, \texttt{output}, etc.
  \item \texttt{question}: The question text.
  \item \texttt{answer}: The final correct answer.
\end{itemize}

\vspace{2pt}
\textbf{2. Core Task}

\begin{enumerate}[leftmargin=*, itemsep=1pt]
  \item First, generate a complete user-facing \textbf{response}: It must reasonably and coherently answer the \texttt{question}, forming a closed explanatory loop.
  \item Then, split the \texttt{response} into multiple continuous sentences/paragraphs (\texttt{sentence}) to form a \texttt{sentence} list; each \texttt{sentence} in the list must have verifiable \texttt{provenance}.
\end{enumerate}

\textbf{Note}: You absolutely do not need, nor are you allowed, to use the following fields (even if they exist in the input):
\begin{itemize}[leftmargin=*, itemsep=1pt]
  \item \texttt{thought\_rethink}
  \item \texttt{thought\_question}
\end{itemize}

\vspace{2pt}
\textbf{3. Output Format (Must be strictly consistent)}

You can only output the following JSON (no additional explanatory text is allowed):

{\footnotesize
\begin{verbatim}
{
  "role": "assistant",
  "content": {
    "response": "...",
    "sentence": [
      {
        "sentence_id": 1,
        "text": "...",
        "provenance": [
          {
            "tool_id": "...",
            "source_text": "...",
            "relation": "..."
          }
        ]
      }
    ]
  }
}
\end{verbatim}
}

\vspace{2pt}
\textbf{4. Response Generation Rules (Ensure completeness and rationality)}

\begin{enumerate}[leftmargin=*, itemsep=1pt]
  \item \textbf{\texttt{response} must answer the \texttt{question}}:
  \begin{itemize}[leftmargin=*, itemsep=1pt]
    \item The \texttt{response} must directly address what the \texttt{question} asks (e.g., ``How many years did the author live?'' must explicitly state the number of years in the \texttt{response}).
    \item The \texttt{response} must contain the literal content of the final \texttt{answer} (e.g., ``55 years``).
  \end{itemize}
  \item \textbf{Restrictions on information sources for the \texttt{response} (Mandatory)}:
  \begin{itemize}[leftmargin=*, itemsep=1pt]
    \item The semantic content of the \texttt{response} can only come from two parts: A) \texttt{context[i].thought} (integrated and rewritten sequentially for i=1..n), and B) \texttt{answer} (used only as the final conclusion).
    \item It is not allowed to introduce new facts outside of \texttt{context.thought} and \texttt{answer}.
    \item It is not allowed to quote or paraphrase fields like \texttt{thought\_rethink}, \texttt{thought\_question}, \texttt{thought\_choose}, or \texttt{thought\_query}.
  \end{itemize}
  \item \textbf{Structural requirements for the \texttt{response} (Mandatory logical closed loop)}:
  The \texttt{response} must contain at least the following logical units (can be expressed in natural language, explicit numbering is not required):
  \begin{itemize}[leftmargin=*, itemsep=1pt]
    \item Identify/determine the subject of the question (e.g., identify the author from the image, or determine key entities).
    \item Obtain key supporting facts (e.g., get birth and death years via OCR/search).
    \item (If needed) Deduce or calculate to get the answer.
    \item Provide the final conclusion (containing the \texttt{answer}).
  \end{itemize}
\end{enumerate}

\vspace{2pt}
\textbf{5. Sentence Generation Rules (Split from response)}

\begin{enumerate}[leftmargin=*, itemsep=1pt]
  \item \textbf{\texttt{sentence} must be split from the \texttt{response}}:
  \begin{itemize}[leftmargin=*, itemsep=1pt]
    \item The \texttt{text} of all sentences in the \texttt{sentence} list, when concatenated by \texttt{sentence\_id} from 1..k (allowed to be connected by spaces or line breaks), must strictly restore the content of the \texttt{response} (semantics and expression must be consistent).
    \item Sentences not present in the \texttt{response} are not allowed to appear in the \texttt{sentence} list.
    \item It is not allowed to have sentences in the \texttt{response} that do not appear in the \texttt{sentence} list.
  \end{itemize}
  \item \textbf{Quantity of \texttt{sentence}s}:
  \begin{itemize}[leftmargin=*, itemsep=1pt]
    \item Determined based on the answer length and the number of tool invocations (i.e., the length of the input \texttt{context}). If the answer is long, the quantity can appropriately exceed the number of tool invocations; if some tools are useless, it can be fewer. Ensure the reasoning chain is complete.
  \end{itemize}
  \item \textbf{\texttt{sentence\_id}}:
  \begin{itemize}[leftmargin=*, itemsep=1pt]
    \item Incrementing from 1, strictly increasing, no skipping numbers.
    \item The order of \texttt{sentence\_id} must be consistent with the order in which the sentences appear in the \texttt{response}.
  \end{itemize}
\end{enumerate}

\vspace{2pt}
\textbf{6. Provenance Rules (Strong constraint, verifiable)}

Each \texttt{sentence} must have at least 1 \texttt{provenance}.

\begin{enumerate}[leftmargin=*, itemsep=1pt]
  \item \textbf{\texttt{tool\_id} naming rules (Mandatory)}:
  \begin{itemize}[leftmargin=*, itemsep=1pt]
    \item \texttt{tool\_id} = \texttt{<context[i].name>\_<i>}
    \item \texttt{i} starts from 1 (the first element of \texttt{context} is 1).
    \item Examples: \texttt{ImageDescription\_1}, \texttt{OCR\_2}, \texttt{GoogleSearch\_3}.
  \end{itemize}
  \item \textbf{\texttt{source\_text} rules (Most important, must be verifiable)}:
  \begin{itemize}[leftmargin=*, itemsep=1pt]
    \item It must contain enough information to support the \texttt{text} of the \texttt{sentence}.
    \item \texttt{provenance.source\_text} must be an exact substring of the corresponding \texttt{context[i].output} (character-by-character match).
    \item Fabricating \texttt{source\_text} is not allowed.
    \item For the ``final answer sentence'', its \texttt{provenance} must quote the tool output fragment that directly supports the \texttt{answer} (usually the last key tool output providing key numbers/facts).
  \end{itemize}
  \item \textbf{\texttt{relation} enumeration (Must choose one of four)}:
  \begin{itemize}[leftmargin=*, itemsep=1pt]
    \item \texttt{Quotation}: Almost copied word-for-word from the original text.
    \item \texttt{Compression}: Compressing and summarizing (still based on the original text).
    \item \texttt{Inference}: Deduce/calculate based on the source (if \texttt{Inference} is used, \texttt{source\_text} must contain the key facts required for deduction, such as years/numbers).
  \end{itemize}
\end{enumerate}

\vspace{2pt}
\textbf{7. Consistency Hard Constraints \& Prohibitions}

\begin{itemize}[leftmargin=*, itemsep=1pt]
  \item The \texttt{response} must look like a real assistant's answer to a user: natural, coherent, and directly answering the question.
  \item The \texttt{response} must contain a chain of explanation, not just list \texttt{thought}s.
  \item The \texttt{sentence} list must be a breakdown of the \texttt{response}, not a mechanical splicing of \texttt{context.thought}.
  \item The \texttt{provenance} of each \texttt{sentence} must be directly related to the content of that \texttt{sentence}; do not provide irrelevant provenance.
  \item \textbf{Strictly Prohibited}: Outputting any text other than JSON; Using a \texttt{tool\_id} that does not exist in \texttt{context}; Any sentence lacking \texttt{provenance}; \texttt{source\_text} not found in the corresponding \texttt{output}; Using undefined categories for \texttt{relation}; Using information from prohibited fields like \texttt{thought\_rethink}/\texttt{thought\_question}; Introducing new facts.
\end{itemize}

\end{promptbox}

\captionof{figure}{Prompt used for the Multi-Tool VQA Provenance Answer Generator. The model generates a coherent response answering the question, splits it into verifiable sentences, and strictly maps each sentence to a provenance source within the tool execution chain.}
\label{fig:multi_tool_vqa_provenance_prompt}

\begin{promptbox}[Prompt for LLM-as-Verifier on Provenance Accuracy]

\sloppy\setlength{\emergencystretch}{1em}

\textbf{1. Task Description}
You need to judge whether the provenance provided by the model is \textbf{completely correct} based on the \textbf{multi-round tool interaction history} and the \textbf{provenance information in the model's final output}, strictly following the 3 rules below.

\vspace{2pt}

\textbf{2. Validation Rules}

\begin{enumerate}[leftmargin=*, itemsep=1pt]

  \item \textbf{Tool and Round Validation}

  \begin{itemize}[leftmargin=*, itemsep=1pt]

    \item \texttt{tool\_id} format: \texttt{ToolName\_Number}

    \item Number = the \textbf{N-th time this tool is called} (incrementing from 1)

    \item Must be completely consistent with the \texttt{tool\_call} order in \texttt{messages}

  \end{itemize}

  \item \textbf{Source Authenticity Validation}

  \begin{itemize}[leftmargin=*, itemsep=1pt]

    \item \texttt{source\_text} must \textbf{completely come from} the original \texttt{tool\_response} result of that specific \texttt{tool\_id} round

    \item Fabricated, spliced, replaced, or non-existent content is not allowed

  \end{itemize}
  \item \textbf{Text and Relation Rationality Validation}
  \newline For each \texttt{sentence\_id}:
  \begin{itemize}[leftmargin=*, itemsep=1pt]

    \item \texttt{Quotation}: The sentence is \textbf{directly copied/extracted} from \texttt{source\_text}

    \item \texttt{Compression}: The sentence is a \textbf{simplification, summary, or compression} of \texttt{source\_text}

    \item \texttt{Inference}: The sentence is derived through \textbf{reasonable deduction} based on \texttt{source\_text}

    \item Must judge: Whether \texttt{text} matches \texttt{source\_text} + \texttt{relation}

  \end{itemize}

\end{enumerate}

\vspace{2pt}

\textbf{3. Input Format}

You will receive a complete JSON containing:

\begin{itemize}[leftmargin=*, itemsep=1pt]

  \item \texttt{images}: Image paths.

  \item \texttt{tools}: List of available tools.

  \item \texttt{messages}: Complete multi-round dialogue of user / model / tool calls / tool returns.

  \item \texttt{solution}: The model's final output, with \texttt{sentence} and \texttt{provenance} tracing.

\end{itemize}

\vspace{2pt}

\textbf{4. Output Requirements}

Please \textbf{only output the structured judgment result}, in the following format:

{\footnotesize

\begin{verbatim}

{
  "overall_correct": true/false,
  "error_details": [
    "Error 1 description",
    "Error 2 description"
  ],
  "sentence_check": [
    {
      "sentence_id": number,
      "tool_id_correct": true/false,
      "source_text_correct": true/false,
      "relation_correct": true/false,
      "sentence_correct": true/false
    }
  ]
}
\end{verbatim}
}

\end{promptbox}

\captionof{figure}{Prompt used to evaluate the correctness of the model's multi-tool provenance. The evaluator verifies tool invocation rounds, source text authenticity, and text-relation rationality, outputting a structured JSON validation result.}

\label{fig:provenance_evaluation_prompt}





\end{document}